\documentclass[journal,transmag]{IEEEtran}
\usepackage{graphicx}
\usepackage{algpseudocode}
\usepackage{algorithm}
\usepackage{amsmath}
\usepackage{epsfig}
\usepackage{subfigure}
\usepackage{epstopdf}
\usepackage{caption}
\newcommand{\argmin}{\operatornamewithlimits{arg\hspace{0.2em} min}}
\usepackage{amssymb,amsthm,url}
\theoremstyle{plain}
\newtheorem{theorem}{Theorem}

\ifCLASSINFOpdf
\else
\fi
\begin{document}
\title{Auto-JacoBin: Auto-encoder Jacobian Binary Hashing}


\author{\IEEEauthorblockN{Xiping Fu,
Brendan McCane,
Steven Mills, 
Michael Albert, and
Lech Szymanski}
\IEEEauthorblockA{Department of Computer Science, University of Otago, Dunedin, New Zealand}
\IEEEauthorblockA{\{xiping, mccane, steven, malbert, lechszym\}@cs.otago.ac.nz}
%\thanks{Manuscript received December 1, 2012; revised August 26, 2015. 
%Corresponding author: Xiping Fu (email: xiping@cs.otago.ac.nz)}
}

\markboth{Journal of \LaTeX\ Class Files,~Vol.~14, No.~8, August~2015}%
{Shell \MakeLowercase{\textit{et al.}}: Bare Demo of IEEEtran.cls for IEEE Transactions on Magnetics Journals}



% for Transactions on Magnetics papers, we must declare the abstract and
% index terms PRIOR to the title within the \IEEEtitleabstractindextext
% IEEEtran command as these need to go into the title area created by
\IEEEtitleabstractindextext{%
\begin{abstract}
Binary codes can be used to speed up nearest neighbor search tasks in large scale data sets as they are efficient for both storage and retrieval. In this paper, we propose a robust auto-encoder model that preserves the geometric relationships of high-dimensional data sets in Hamming space. This is done by considering  a noise-removing function in a region surrounding the manifold where the training data points lie. This function is defined with the property that it projects the data points near the manifold into the manifold wisely, and we approximate this function by its first order approximation. Experimental results show that the proposed method achieves better than state-of-the-art results on three large scale high dimensional data sets.
\end{abstract}

}

\maketitle

\IEEEdisplaynontitleabstractindextext

% \fi
\IEEEpeerreviewmaketitle

\section{Introduction}
Finding nearest neighbor points is a computational task which arises frequently in machine learning, information retrieval and computer vision. Naive search is infeasible for massive real-word datasets since it takes linear time complexity to retrieve the nearest point in the dataset.  For low dimensional datasets, the nearest neighbor searching problem can be addressed by efficient and effective tree based approaches such as {\it KD-tree} \cite{bentley1975multidimensional} and {\it R-tree} \cite{guttman1984r}. Due to the curse of dimensionality, when it comes to the high dimensional scenario, tree based approaches deteriorate abruptly and can become worse than linear search complexity \cite{weber1998quantitative}. Recent years have witnessed a surge of research interest in hashing methods which return approximate nearest neighbor ({\it ANN}) data points. It has been shown that binary hashing based {\it ANN}, which indexes data points into binary codes and retrieves data points in Hamming space, often has good enough performance for many real world applications. For example, hashing has been used in large scale medical image searching \cite{yu2013large}, image matching for 3D reconstruction \cite{cheng2014fast}, collaborative filtering for recommender systems \cite{shrivastava2014asymmetric} and approximating the {\it SVM} kernel \cite{muhash2014Hash}.

When high dimensional data points are indexed into compact binary codes, information loss is the main concern. Most previous work focuses on modeling the obtained binary codes optimally, i.e., attempting to ensure that the geometric relationships among data points in the new space are similar to those in the original space. Locality sensitive hashing ({\it LSH}) \cite{indyk1998approximate} was an early exploration of hashing to encode data points into binary codes. In {\it LSH}, each binary code is obtained by a randomly generated projection matrix. The randomness of the projection matrix insures that similarities between data points are inherited by the binary codes if sufficiently many bits are used to encode the data points. In more recent research, {\it LSH} has been used for the nearest neighbor searching problem for the $L_p$ norm \cite{datar2004locality}, the learned Mahalanobis metric \cite{jain2008fast}, the reproducing kernel Hilbert space \cite{kulis2009kernelized} and for maximising inner products \cite{shrivastava2014asymmetric}. All of these algorithms belong to the category of data independent hashing algorithms where the binary codes are obtained from some random matrices. 

Other algorithms compute hashes based on the data being indexed. Since the bits are used effectively, compact binary codes obtained from data dependent methods often have better performance when they are used for retrieving nearest neighbor points. One of the seminal approaches in this category is {\it Spectral Hashing} \cite{weiss2009spectral}, which models neighborhood relationships from the original space in Hamming space. That is, the binary codes in the Hamming space should be close if their corresponding data points in the original space are close. This hashing approach has been extensively developed in recent years \cite{wang2010sequential, liu2011hashing, liu2014discrete, xu2013harmonious}. 
In some data sets, label information is available for supervised methods. For example, in both supervised hashing \cite{liu2012supervised} and minimal loss hashing \cite{norouzi2011minimal},  the similarity matrix is used to construct the loss function. Alternatively, binary codes may be designed by minimizing the quantization error between the binary codes and the features obtained from original data set \cite{gong2011iterative, he2013k, norouzi2013cartesian, fu2014}.

Motivated by the success that auto-encoder models have had in preserving geometric information of high dimensional data sets \cite{van2009dimensionality, hinton2006reducing}, we propose to use auto-encoders to learn binary codes for hashing. Our optimisation model is constructed from both the auto-encoder model and the optimal binary code's perspectives. 

From the auto-encoder model, we use a three layer network to keep geometric information consistent with the high dimensional data points. The common approach for auto-encoder models is to assume that the data points are sampled from some manifold. The task is to find a set of parameters for the neural network such that the data points from the manifold are reconstructible. This means that the gap between the output data points and the input training data points is as small as possible. 
Points drawn from a domain around the manifold can be thought of as noisy data points. It is one of the features of the neural network model that it can process such points. Indeed, properly modeling the functions in this larger domain gives better feature learning capability for the model.
Motivated by some recent efforts which aim to remove noise during the neural network forward propagation \cite{vincent2010stacked, rifai2011contractive}, we define the optimal noise removing function directly and investigate its first order approximation which is used as a component in our later optimization model.
We believe that the first order information from the optimal noise removing function makes the learned auto-encoder model robust, and enables the discovery of local relationships between the data points consistent with geometric information. 

From the optimal binary code perspective, our final interest is to obtain binary codes. Since the auto-encoder model does not provide these codes directly, we make use of the common constraint that minimizes the gap between the learned features and the ideal binary codes.
We use an approximated 1-norm to model the gap minimizing problem aiming for a better distribution in Hamming space for the binary code. 

Our main contributions are as follows: 
\begin{itemize}
\item A novel noise removing function. We investigate a function which has a noise absorbing property, and find that its Jacobian matrix has an intimate relationship with the tangent space in the local region.

\item A robust constraint that encourages the binary codes to have an optimal distribution in Hamming space. 

\item  A novel hashing algorithm. The auto-encoder model is adapted for generating effective binary codes as well as preserving geometric relationships of the original real-world data sets. 
\end{itemize}

\section{Background}
\subsection{Notation}
\begin{figure}[ht]
\centering
\includegraphics[width=8.0cm]{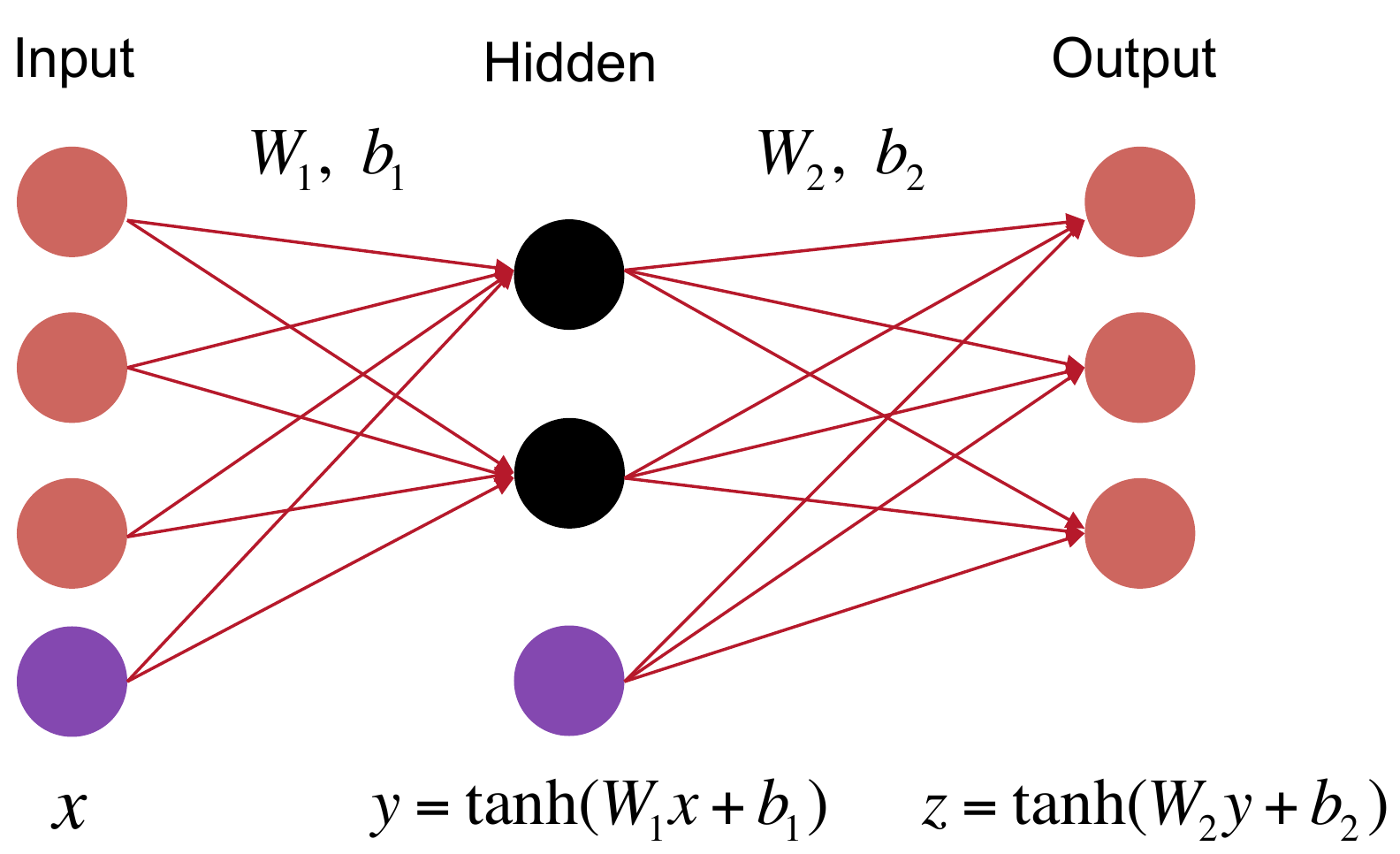}
\caption[The Auto-encoder model used in our proposed method.]{The Auto-encoder model used in our proposed method. The hidden layer is colored in black, the biases in purple. The forward calculation is shown under the corresponding layer.}
\label{fig:tree}
\end{figure}

Suppose $X=[x_1, x_2, \cdots, x_N] \in \mathcal{R}^{D\times N}$ are the training data points, we denote the underlying distribution of the training data as the manifold $\mathcal{M}$, and, for each $x_i\in \mathcal{M}$, $T_i\in \mathcal{R}^{D\times d_\mathcal{M}}$ is a basis for the tangent space at $x_i$. We use the three layer auto-encoder model to ensure the second layer (hidden layer) captures the geometric relationships in the training data set. Fig.~\ref{fig:tree} displays the three layer neural network used in our method. Denote $W_1\in \mathcal{R}^{d\times D}$ and $W_2\in \mathcal{R}^{D\times d}$ as the weight matrices which connect the neural network from the first layer (input layer) to second layer and the second to third layer (output layer),  and the biases in the second and third layers are denoted as $b_1$ and $b_2$. The features from the second layer are denoted as $Y=[y_1, y_2, \cdots, y_N] \in \mathcal{R}^{d \times N}$ where $y_i=\tanh(W_1x_i+b_1)$ and the features in the third layer are denoted as $Z=[z_1, z_2, \cdots, z_N] \in \mathcal{R}^{D\times N}$ where $z_i=\tanh(W_2y_i+b_2)$. The main assumption of auto-encoder model is that input and output should be as similar as possible. Thus the object of the auto-encoder model is to find a set of parameters for the neural network such that 
\begin{equation}\label{eq:auto}
\mathcal{C}_1(W_1, W_2, b_1, b_2)=\sum_{i=1}^{N}||z_i-x_i||^2
\end{equation}
is minimized. The final binary code is obtained by $B=\text{sign}(W_1X)\in\{-1, 1\}^{d \times N}$. 

\subsection{Unified view of auto-encoders}
For noise-free data, we expect that data points can be reconstructed exactly. Thus, it can be directly done by minimizing $\mathcal{C}_1(W_1, W_2, b_1, b_2)$. For noisy data, consider a data point $x$ out of the manifold $\mathcal{M}$ and $z$ is the corresponding output, minimising the distance between $x$ and $z$ might deteriorate the final performance of the auto-encoder model since the ideal output should be the noiseless data point $m$.
In order to account for both the noise-free and noisy scenarios, we reformulate the optimization objective (Equation \eqref{eq:auto}) of the auto-encoder model in a larger region in a unified way: denote $f$ as a function which is defined around the manifold $\mathcal{M}$ and the restriction of $f$ to $\mathcal{M}$ is the identity, i.e., $f(x)=x$ for $\forall x \in \mathcal{M}$. Thus the unified optimization objective is to find a set of parameters for the neural network such that 
\begin{equation}\label{eq:autore}
\mathcal{C}_2(W_1, W_2, b_1, b_2)=\sum_{i=1}^{N}||z_i-f(x_i)||^2
\end{equation}
Note that for noise-free data, Equation \eqref{eq:auto} is equivalent to Equation \eqref{eq:autore}, while for noisy data, the condition $f(x_i)=x_i$ might not be satisfied and we have to define $f$ appropriately. 

\subsection{Related work}
Recent efforts for dealing with noisy data are the denoising auto-encoders ({\it DAEs}) \cite{vincent2010stacked} and contractive auto-encoders ({\it CAEs}) \cite{rifai2011contractive}. The main starting points of these algorithms are that the learned features in the hidden layer keep the important intrinsic structure of the original data set while unimportant information, such as the noise, is discarded as much as possible. In {\it DAEs}, noise is manually added to the training data points; since we know the correspondence between the noisy and clean versions, a constraint is introduced to minimize the gap between these two versions according to some loss function. In {\it CAEs}, the Jacobian norm of the function, which maps the input data points to the hidden layer, is minimized for a contractive effect. In the extreme case, {\it CAEs} may contract all of the data points in the original space to a single point. This constraint is used to discard noise, but the distance between the input and output points is also considered. In the balance of these two constraints, data on the manifold will be unchanged and data outside the manifold will contract to the manifold.

Both {\it DAEs} and {\it CAEs} focus on training a model such that noisy data are projected onto the manifold and non-noisy data are left alone.  When the data points  have some noise, i.e., they are distributed around the manifold, the output and hidden layers should keep the important information and at the same time be robust to some degree of noise. 
This process is done implicitly with both approaches. {\it DAEs} introduce artificial noise, and assume that eliminating artificial noise will also help to eliminate real noise, i.e., the ideal noise data point is projected to the original data point through the forward propagation of the learned neural network. For {\it CAEs}, the noise discarding process is done through balancing geometric consistency and the contractive property by minimizing the norm of the Jacobian matrix. This motivates us to explore an explicit function which has the ideal noise absorbing property. 

\begin{figure}[ht!]
\centering
\includegraphics[width=4.2cm]{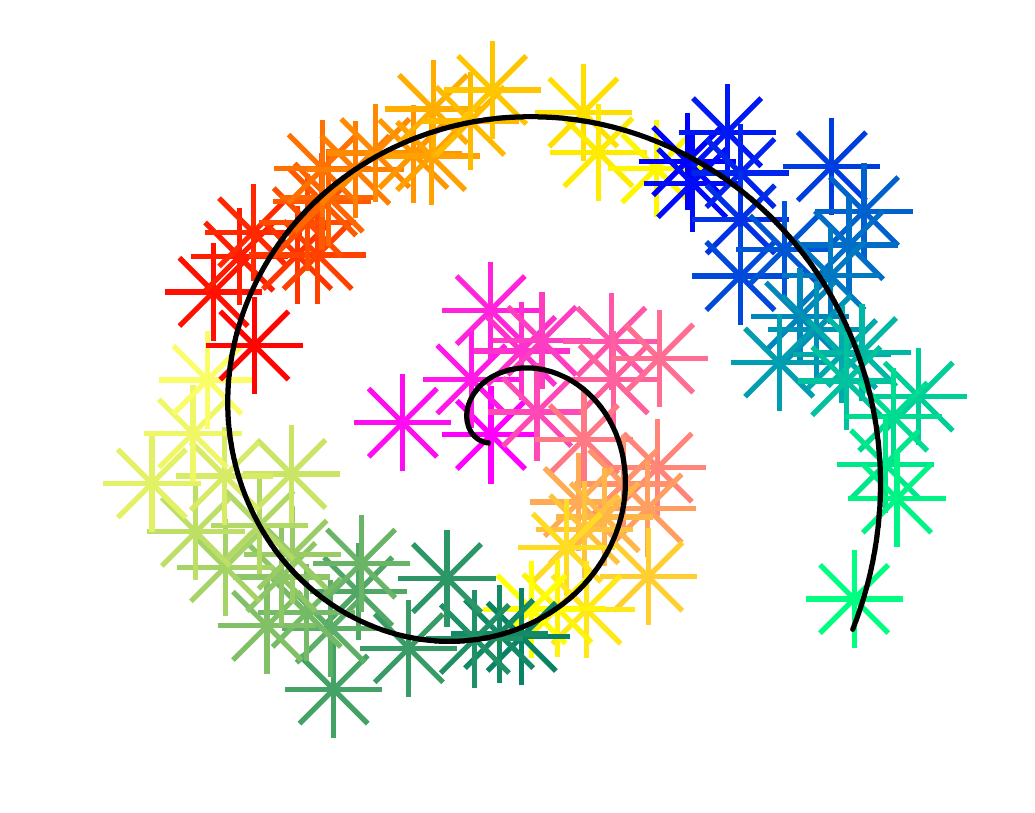}
\includegraphics[width=4.2cm]{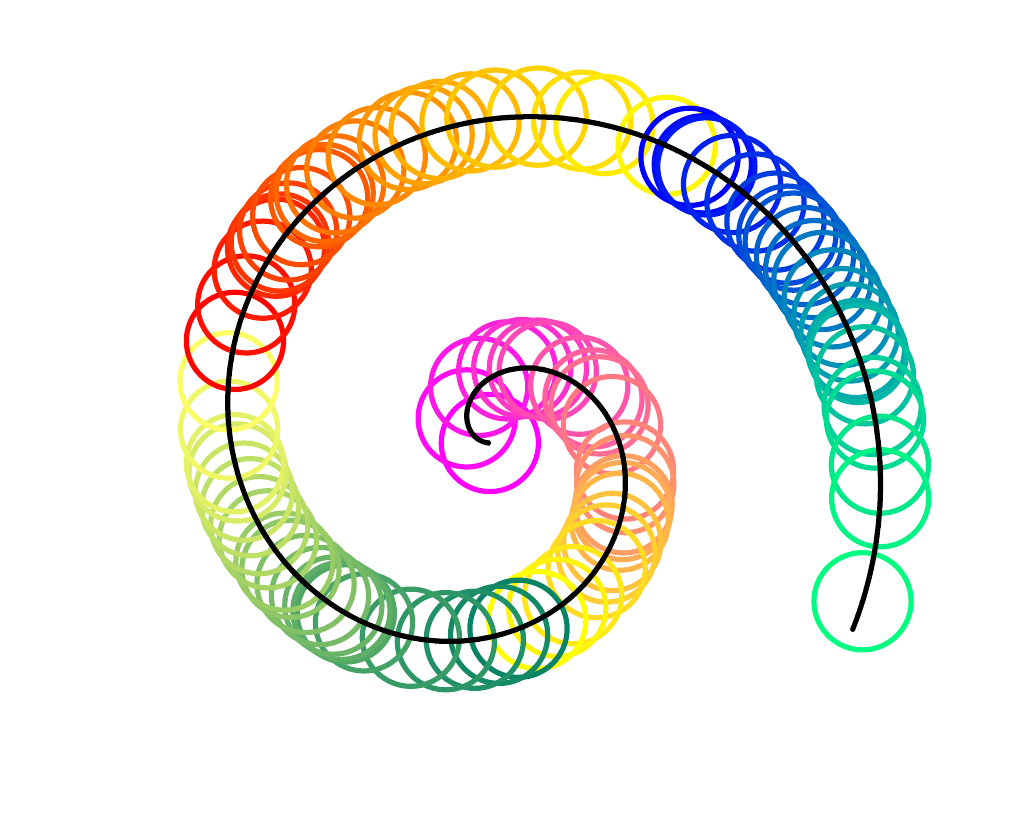}
\caption[The ideal function which projects data point near to the manifold onto its closest data point on the manifold.]{The ideal function which projects data point near to the manifold onto its closest data point on the manifold. In the left figure, the colored points, which are sampled around the black curved manifold, are the input data points. Here different data points are distinguished with different colors. The right figure shows the corresponding output data points of the function. Here each circle represents an output data point and corresponds to the star with the same color in left figure. }
\label{fig:denoisingFunction}
\end{figure}

Fig.~\ref{fig:denoisingFunction} displays one of the functions we are going to explore. Notice that each data point around the manifold is exactly projected to the nearest point on the manifold, while {\it DAEs} and {\it CAEs} do not have this kind of guarantee  albeit they are designed to discard noise.

\section{Algorithm}
\subsection{Motivation}
The binary codes we seek have some intersections with general feature learning in auto-encoder models. On one hand, the features of both approaches are from the hidden layer with the aim that features in the hidden layer have the same amount of information as in the original data set.   On the other hand, previous auto-encoder models focus on obtaining discriminative features since their main purpose in learning these features is to train a classifier, and the auto-encoder model is viewed as a building block for a deep neural network. For hashing, we hope that the features in hidden layers are geometrically consistent with the original space, i.e., the relative order information between the data points is preserved, and the hidden features should be close to the binary codes under some metric.

Notice that both {\it CAEs} and {\it DAEs} try to project data points around the manifold to data points on the manifold. This motivates us to define a function $f$ with this property directly, and the data points around the manifold can be viewed as noisy data points, i.e., they are generated by data points from the manifold plus some noise. Without any prior information we assume that each noisy data point is generated from the closest point on the manifold. Formally, we seek a function, $f$, such that
\begin{equation}
f(x)=\argmin_{m\in \mathcal{M}}||x-m||_2^2.
\end{equation}
The object of training is to find a set of parameters for the network such that $\sum_{i=1}^{N}||z_i-f(x_i)||^2$ is minimized.

The function $f$ is only valid in some proper regions. This is because for data points outside the manifold, their closest point on the manifold might not be unique. In practice, we assume that the noise is constrained to a limited region around the manifold and therefore $f$ is valid most of the time. 
Minimizing the distance between $x$ and $z$ can be viewed as the $0^{th}$-order approximation of the function $f$. More information about $f$ can be captured by using higher order approximations and the Jacobian matrix of $f$ is intimately related with the tangent space of $\mathcal{M}$:

\begin{theorem}\label{mcase}
Suppose $\mathcal{M}$ is a $d$ dimensional compact smooth submanifold in $\mathcal{R}^D$, $f$ is a function defined as $f(x)=\argmin_{m\in \mathcal{M}}||x-m||_2^2$,  $\forall x \in \mathcal{R}^D$ . For each $m$ in $\mathcal{M}$, let $T_m\in \mathcal{R}^{D\times d}$ be the local normal basis of the tangent space to $\mathcal{M}$ at $m$. Then, the Jacobian matrix of $f$ at $m$ is $T_mT_m'$.
(The detailed proof is provided in Appendix I.)
\end{theorem}

If we view the tangent space as a point in a Grassmannian manifold, Theorem \ref{mcase} tells us that the Jacobian matrix at $m\in \mathcal{M}$ is exactly the point where the Grassmannian point $T_m$ is embedded in $\mathcal{R}^{D\times D}$.
Assuming the training data points are sampled from the manifold $\mathcal{M}$, for each  point $x_i$, we estimate its tangent space $T_i$ by the local {\it PCA} technique. So now, the manifold $\mathcal{M}$ is approximated by tangent patches at the training data points. On the other hand, when parameters of the three layer neural network are fixed, the output features can be viewed as a function of the input features. Thus, the Jacobian matrix of the function obtained from the neural network can be analytically expressed. Specifically, for the data point $x_i$, the Jacobian matrix $J_i$ can be calculated by chain rule on the network:
\begin{equation}J_i=W_1'\left( W_2'\odot\left(1-z_2^2\right)\left(1-z_3^2\right)'\right), \end{equation}
where $\odot$ is the point-wise product operator between two matrices, $z_2$ and $z_3$ are the corresponding features in second and third layer of the network, and $(\cdot)^2$ is the element-wise square of the entries of a vector. 
For the $i^{\mbox{\scriptsize th}}$ data point, we have $J_i$ based on the parameters of the network, and we also have an approximation for $T_i$ based on the data. Therefore, we propose to minimize the distance between $J_i$ and $T_iT_i'$ in our optimization model.

Since the features in the hidden layer range from $-1$ to $1$, and binary codes are what we really need, we use the metric $||YY'-NI||_1$ to constrain the features in the hidden layer. The reason for this metric is that since the ideal elements in Y are either $1$ or $-1$, the ideal diagonal position of $YY'$ should be $N$ exactly, and the non-diagonal position in $YY'$ should be $0$ to favour uncorrelated binary codes.  The 1-norm, $||\cdot ||_1$, is calculated by summing the absolute value of the matrix's elements. Since the absolute value function is non-differentiable at $0$, we approximate it by introducing a constant value $\epsilon=0.0001$, giving $||a||_1 \approx \sqrt[]{a^2 + \epsilon}$. We denote this approximate 1-norm as $||\cdot||_1^\epsilon$. 

Finally, the optimization objective is
\begin{align}
&\min \mathcal{C}_3(W_1,W_2,b_1,b_2)  \label{eq:autojacobinObject}\\
=&\sum_{i=1}^{N}(||x_i-z_i||_F^2+||J_i-T_iT_i'||_F^2) +\alpha||YY'-NI||_1^\epsilon,\notag
\end{align}
where $N$ is the number of training data points and $\alpha$ is a weight parameter balancing the optimal binary codes and the geometric relationships from the training data set. We haven't introduced a weight for the Jacobian term to keep it consistent with the Taylor series expansion.

\begin{figure}[ht!]
\centering
\includegraphics[width=8cm]{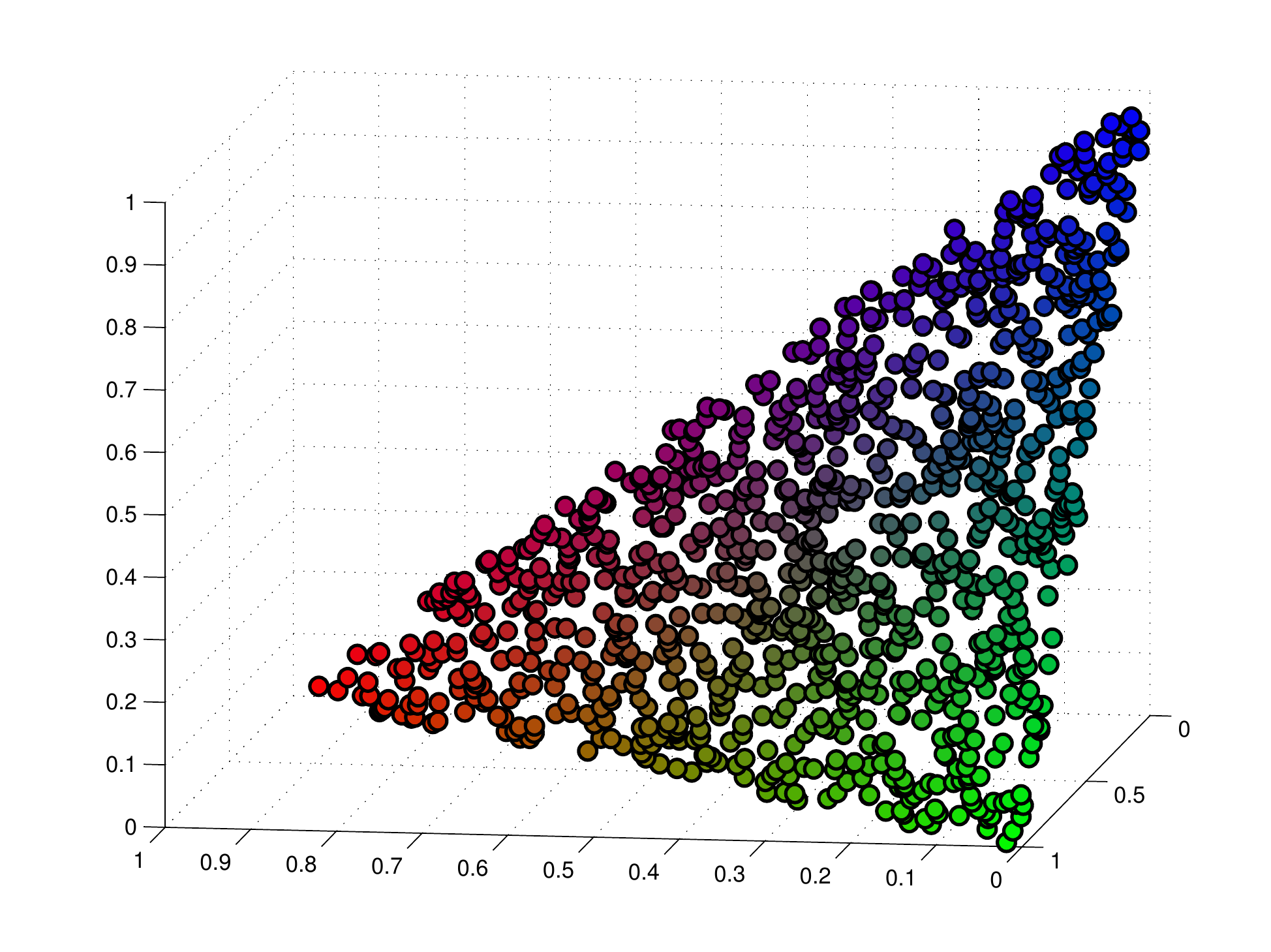}\hspace{-1cm}
\includegraphics[width=8cm]{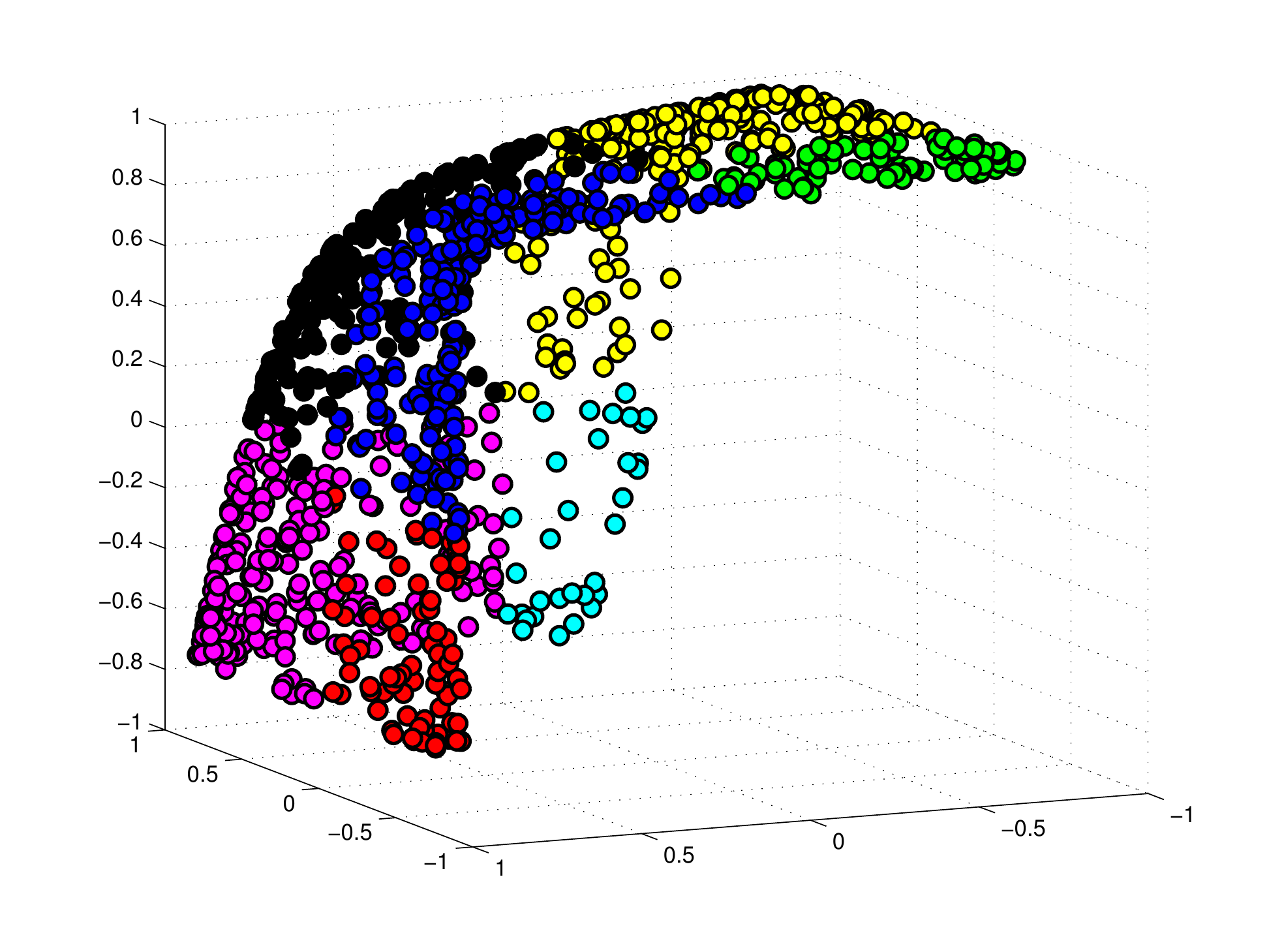}
\caption[Visualizing of the warping.]{Visualizing of the warping. The top picture shows the training data set, which consists of $1,000$ data points sampled from $\{(x_1, x_2, x_3) ~|~ x_1+x_2+x_3=1~~ \mbox{and}~~ x_i>0\}$. The bottom picture shows the hidden features. The color of the data points are distinguished according to their positions in different quadrants. The neural network has the effect to warp the manifold of the data set into the surface of the cube in the $3$D space. }
\label{fig:ToyAutoJacobin}
\end{figure}

Fig.~\ref{fig:ToyAutoJacobin} shows a toy example for the visualization of the effects of the two optimization components. The data points are randomly sampled from a plane $\{(x_1, x_2, x_3) ~|~ x_1+x_2+x_3=1~~ \mbox{and}~~ x_i>0\}$. The top picture in Fig.~\ref{fig:ToyAutoJacobin} shows the training data points in the 3D Euclidean space. After optimization, the hidden features are plotted in the bottom picture. Since we want to encode the data points into binary vectors, the data points are distinguished according to their positions in different quadrants. The visualization of the hidden features fully explains the motivation of the proposed optimization model. Firstly, the robust auto-encoder is used to preserve the geometric information. For example, the transform from the training data points to the hidden features behaves like warping a piece of paper and the relative (distance) order information in local region is kept.  Secondly,  the constraint on the hidden features has the effect to disperse the hidden features to the vertices of the cube. Note that we only use three bits to encode the training data set which is sampled from a 2D plane, the number of different binary codes is at most 7.  From the color in the bottom picture, we can see that the data points are encoded with 7 different binary codes according to their positions in different quadrants. Thus, the constraint makes us use the binary codes fully. In all,  the proposed optimization model is capable of maintaining geometric information as well as learning optimal binary codes.

\subsection{Optimization}
To optimize the objective function $C$, we update the parameters by a mini-batch stochastic gradient descent method \cite{bengio2012practical}.  
Since Equation \eqref{eq:autojacobinObject} is non-convex, using mini-batches achieves a better solution in general, which motivates the use of the stochastic gradient descent method for learning the parameters. The maximum number of epoch is set to be $I_{max}$. In each epoch, the training data is randomly divided into $m$ mini-batches and the parameters are updated with each mini-batch. For updating the parameters, we have to calculate the gradients: $\frac{\partial C}{\partial W_1}$, $\frac{\partial C}{\partial W_2}$, $\frac{\partial C}{\partial b_1}$ and $\frac{\partial C}{\partial b_2}$ respectively. The detailed gradient calculation is provided in Appendix II. Denote $\theta_i$ as the vector of parameters $W_1$, $W_2$, $b_1$ and $b_2$ at the $i^{\mbox{\scriptsize th}}$ iteration and $G_i$ as the gradient of $C$ at the $i^{\mbox{\scriptsize th}}$ iteration. The step length $\lambda_i$ is found by satisfying the Wolfe conditions \cite{schmidt2012minfunc}.

Since $\tanh$ is used in the neural network, the data points have to be normalized properly. In our experiments, we scale the data points by the inverse of the maximum norm from the training data set, and then scale the data points by $0.8$ again as the training data points might not fully cover the distribution of the base and query data sets.  For initializing the parameters, suppose $\mu$ is the mean of the training data points, $W_1$ is initialized by the {\it PCA} projection times a random rotation matrix, and $W_2$, $b_1$ and $b_2$ are initialized as $W_1'$, $-W_1\mu$ and $\mu$ respectively. The reason that we use this set of initial parameters is that $\tanh$ can be approximated by a linear function around the origin by Taylor expansion method, and this set of parameters is optimal for reconstruction error when a linear function is used in the neural network. For estimating the tangent plane $T_{x_i}(\mathcal{M})$, we retrieve the $(D+d)$ nearest data points for each $x_i$, and take the {\it PCA} projection which preserves $98\%$ of the energy or the most energetic $d$ dimensions, whichever is smaller. Algorithm~\ref{alg:AutoHashing} is a summary of the proposed training process.

\begin{algorithm}[ht]
  \caption{Auto-JacoBin}
  \label{alg:AutoHashing}
  \begin{algorithmic}[1]
   \State {\bfseries Input:} Training data points $x_1, x_2, \cdots, x_N$, the maximum number of iteration $I_{max}$, and the number of batches $m$.
   \State {\bfseries Output:} The parameters $W_1, W_2, b_1$ and $ b_2$.
   
   \State Initialize $W_1, W_2, b_1, b_2$, and then vectorize them into a vector $\theta$. 
   \For{$i=1$ {\bfseries to} $N$}
    \State Estimate the tangent space $T_{x_i}(\mathcal{M})$.                 
   \EndFor   
   
   \For{$i=1$ {\bf to} $I_{max}$}
   \State Randomly permute the order of the training data set, 
   
   ~and divided the training data into $m$ subsets.
   \For{$j=1$ {\bfseries to} $m$}
    \State Calculate the gradient $G_{ij}$ of the current $j^{\mbox{\scriptsize th}}$ subset.
    \State Find the optimal step length $\lambda_{ij}$ such that the 
    
    ~~~~~Wolfe conditions are satisfied.
    \State  Update the parameter $\theta$: $$\theta \leftarrow \theta-\lambda_{ij} G_{ij}$$    
   \EndFor
   \EndFor
    \State Reshape the vector $\theta$ into parameters $W_1, W_2, b_1$ and $b_2$.
 \end{algorithmic}
\end{algorithm}

\subsection{Computational complexity}
For the tangent plane estimation stage, the $K$ nearest neighbors are located by brute-force which takes $\Theta(N^2K)$ computation time and the spectral decomposition takes $\Theta(ND^3)$.
In each iteration, calculating $Y$ and $Z$ takes $ \Theta(NDd)$ time, and both the Jacobian matrices and the gradient calculation take $ \Theta(ND^2d)$. Since we use a line search method to find the step length, it might take multiple evaluations to satisfy the Wolfe conditions. It is too computationally and memory expensive to use all training data points at each optimisation step, and therefore we approximate by using a batch size of $1,000$ points. So $m$ is $\frac{N}{1000}$ in our experiments. 

The proposed method takes relatively high computation during the training stage. In our experiment, for SIFT1M data set (N=10,000, D=128 and d=64), it takes about 2 minutes to estimate the tangent planes and 6 minutes to learn the parameters of the neural network, while the training times for {\it NOKMeans}, {\it OKMeans}, {\it ITQ} and {\it Spectral Hashing} are 15.18, 3.67, 1.79 and 1.229 seconds respectively. We should note that the offline training stage is only done once, and at runtime the method has the same encoding and retrieving time as other hashing methods. 

\section{Experiments}
To evaluate the performance of Auto-JacoBin, its performance is compared with several state-of-the-art hashing algorithms including {\it LSH} \cite{indyk1998approximate}, {\it Spectral Hashing} \cite{weiss2009spectral}, {\it ITQ} \cite{gong2011iterative}, {\it OKMeans} \cite{norouzi2013cartesian}, and {\it NOKmeans} \cite{fu2014}. We also compare variants of Auto-JacoBin including different constraints and different optimization methods. We conduct experiments on two benchmark image data sets and two local feature data sets. Two data sets use global image features: 960D GIST features~\cite{oliva2001modeling} in the GIST1M \cite{jegou2011product} set; and 128D wavelet texture feature \cite{manjunath1996texture} in the NUS-WIDE \cite{chua2009nus} set. We also use two local feature data sets, SIFT1M~\cite{jegou2011product} and SIFT10M \cite{fu2014}. In SIFT1M, each data point is a 128D local SIFT feature~\cite{lowe2004distinctive} and extracted from Flickr images and INRIA Holidays images \cite{jegou2010improving}. 

\begin{figure}[htbp!]
\centering
\includegraphics[width=8.0cm]{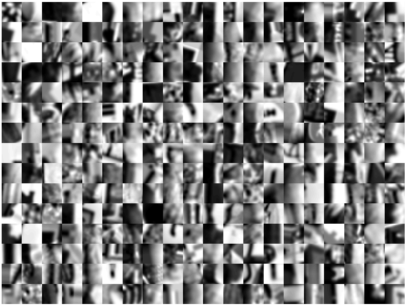}
\caption[Image patches associated with randomly chosen features from SIFT10M data set.]{Image patches associated with randomly chosen features from SIFT10M data set. Each of the features is associated with a patch which is used for visualization purposes. The keypoint detection method is often used to localize the interest regions, and then SIFT features are calculated.  Due to the fact that boundaries or corners are important for representing images, from the figure, we can see that most of the patches correspond to this kind of information.    }
\label{fig:SIFT10MRandomPatches}
\end{figure}

SIFT10M is another SIFT feature data set, and it is extracted from Caltech-256 \cite{Griffin2007}. For each SIFT feature in SIFT10M data set, we have the corresponding image patch which provides a kind of `visualization' of the corresponding SIFT feature and helps us analyze the performance of different hashing methods. Fig.~\ref{fig:SIFT10MRandomPatches}  shows some random collection of these patches. From the figure, we can see the corresponding patches of the SIFT features in this data set. 

In all four data sets, $10,000$ points are chosen for training. For GIST1M and SIFT1M, we use the provided query and base data sets. For SIFT10M, the base set consists of $10,000,000$ points and it has the same number of points in the training and query sets as SIFT1M. For NUS-WIDE, we choose $10,000$ points as the query points, and the remaining as the base points. Summary information about these four data sets is given in Table~\ref{table:datasets}.

\begin{table}[htb]
\centering
 \begin{tabular}{|c|c|c|c|c|}
       \hline
       {\bf Data set}  &{\bf dimension}& {\bf base set} & {\bf training set} & {\bf query set} \\
       \hline
       {\bf SIFT1M} & 128&$10^6$  & $10^4$ & $10^4$ \\
       \hline
       {\bf SIFT10M} & 128& $10^7$ & $10^4$ & $10^4$ \\
       \hline
       {\bf GIST1M} & 960& $10^6$ & $10^4$ & $10^3$ \\
       \hline
       {\bf NUS-WIDE} & 128&$294,648$ & $10^4$ & $10^4$ \\
       \hline
 \end{tabular}
 \caption[Data sets which are used for evaluating different approximate nearest search algorithms.]{Data sets which are used for evaluating different approximate nearest search algorithms. SIFT1M, SIFT1B and GIST1M are three benchmark data sets for testing the performance of hashing methods. SIFT10M is prepared for the purpose of visualization as well as a medium size data set between SIFT1M and SIFT1B.}
 \label{table:datasets}
\end{table}

\subsection{Performance measurements}
\label{subsec:meansure}
\begin{figure*}[htbp!]
\vskip -0.5cm
\centering
\includegraphics[width=18.0cm,height=17.0cm]{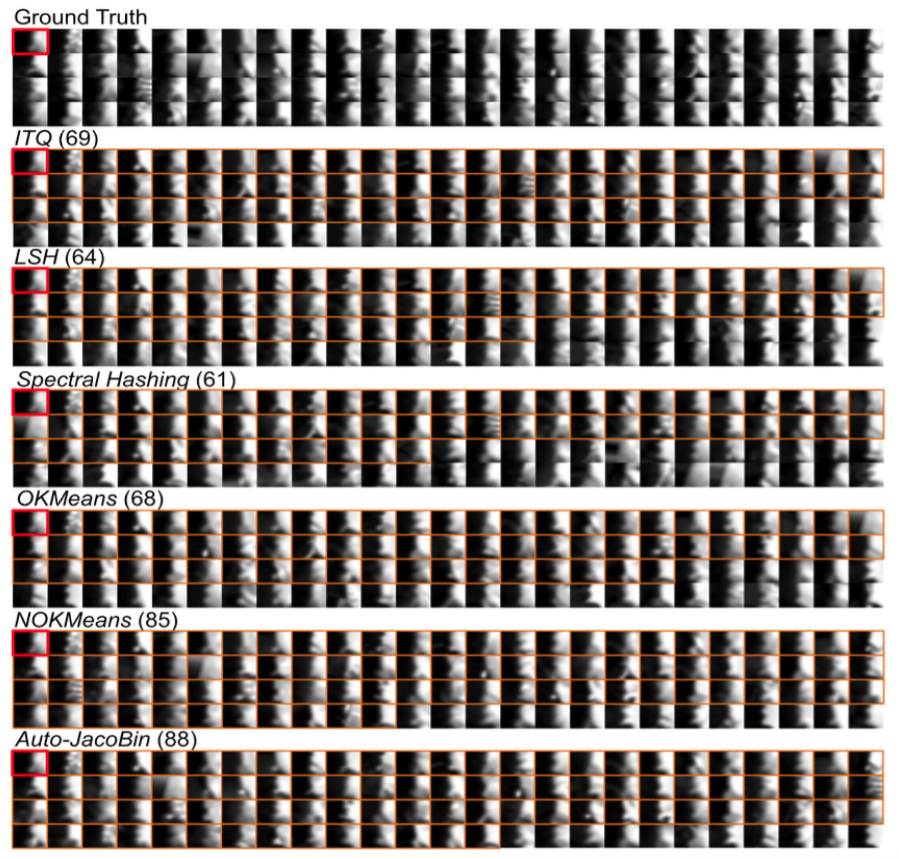}
\caption[Query results of SIFT10M data set.]{Query results of SIFT10M data set. The top picture shows the patches of the query point and its true $99$ nearest neighbor data points in sense of Euclidean distance. The other pictures show the retrieval results of different hashing methods. Each data point is encoded into $128$D binary codes, $0.01\%$ of base data is retrieved, and from among those the $99$ nearest points in sense of the Euclidean distance are returned. The patches with orange border are among the true $99$ nearest neighbor data points, and the number of blue squares is shown in the brackets. }
\label{fig:sift10Mqueryindex1}
\end{figure*}

To evaluate the performance of different hashing methods, we first investigate how hashing methods are used for {\it ANN} problems. We take SIFT10M as an example since the corresponding patches provide the visualization of the retrieval results. Fig.~\ref{fig:sift10Mqueryindex1} shows an example of finding the nearest neighbor data points for the query point. The top picture shows the patches of the query point and its true $99$ nearest neighbor points among the $10,000,000$ data points. Here the patch of the query point is highlighted by a red border, and the patches of the true $99$ nearest neighbor points are ordered from best to worst along the rows. From top picture, we can see that SIFT features provide a faithful representation of image patches since the corresponding patches of the neighboring features are visually similar.

For each hashing method, we use $128$ bits to encode each SIFT feature and retrieve the $1,000$ nearest neighbor  data points in sense of the Hamming distance. Here we should note that the $1,000$ data points might not be exactly the same as the true $1,000$ nearest neighbor  data points in sense of the Euclidean distance and we only retrieve $0.01\%$ of the base data set. Then we retrieve the $99$ nearest neighbor data points among the $1,000$ selected points in sense of the Euclidean distance, and the $99$ data points are placed accordingly in each subfigure. If the data point is among the true $99$ nearest neighbor points of the query in sense of the Euclidean distance, we highlight it by an orange border, otherwise the patch is plotted without any border.

From Fig.~\ref{fig:sift10Mqueryindex1}, we can see that although the hashing methods return approximate nearest neighbor data points, i.e., the retrieved 99 points are not exactly the same as the true 99 nearest data points, the returned data points are very close to the query data points since their corresponding patches have similar visual appearance. Thus, this is the main reason why {\it ANN} is useful in practice. On the other hand, although the hashing methods return satisfactory results, we hope it can return the true nearest neighbor data points in sense of Euclidean distance as frequently as possible since this would guarantee the reliability of image retrieval or feature matching. Thus, the percentage of the true nearest neighbor data points retrieved is an important indicator for evaluating the performance of different hashing methods. From the number of orange borders in Fig.~\ref{fig:sift10Mqueryindex1}, we can see that most of the true nearest neighbor data points are retrieved when only $0.01\%$ of the base data set is considered. According to this indication, {\it Auto-JaboBin} has the leading performance for this query point since it has the highest percentage for retrieving the true 99 nearest neighbor data points.

Based on the desirable features of an {\it ANN} method in practice as outlined above, we adopt recall as the main indicator for evaluating retrieval performance. For the $j^{\mbox{\scriptsize th}}$ query data point, denote $TE_j$ as the set of $k$ true Euclidean nearest neighbor points and $TH_{ij}$ as the $i$ data points retrieved in Hamming space, the recall is defined as the proportion of true nearest neighbor points retrieved and $\text{Recall@}i$ is defined as the average recall performance of all query points. Formally, it is defined as
\begin{equation}\label{eq:RecallNeq}
 \text{Recall@}i=\frac{1}{N}\sum_{j=1}^N\frac{|TE_j \cap TH_{ij}|}{|TE_j|}
  \end{equation}
In order to see the performance of the hashing methods when different number of data points in Hamming space are retrieved, $i$ is varied from $1$ to $K=10,000$ in the following experiments. 

For evaluating the overall retrieval performance, we use the m-Recall measure which averages $\text{Recall@}i$ when the number of data points retrieved in Hamming space is up to $K$ and it is defined as:
 \begin{equation}\label{eq:mREcalleq}
 \mbox{m-Recall}=\frac{\sum_{i=1}^{K} \text{Recall@}i}{K}.
 \end{equation}

Another related indicator for measuring the retrieval performance is precision. Actually, the only difference of the definitions between recall and precision is the denominator, which is the number of true nearest data  points for recall and the number of retrieved data points in precision. From the definitions, we can see that recall emphasizes the percentage of true nearest neighbor points retrieved, while precision emphasizes the percentage of true nearest neighbor points among the retrieved data points. As discussed previously $k$ is much smaller than $K$ for the {\it ANN} problem, the precision will be almost 0 due to the high value of $K$ regardless of whether the true nearest neighbor points are retrieved or not. For massive data sets, we are most interested in the recall performance when only a small proportion of the database is retrieved. From the definition of m-Recall, we can see that it reflects overall performance for this scenario. 

\subsection{Parameter selection}
\begin{figure}[ht!]
\centering
\includegraphics[width=8cm]{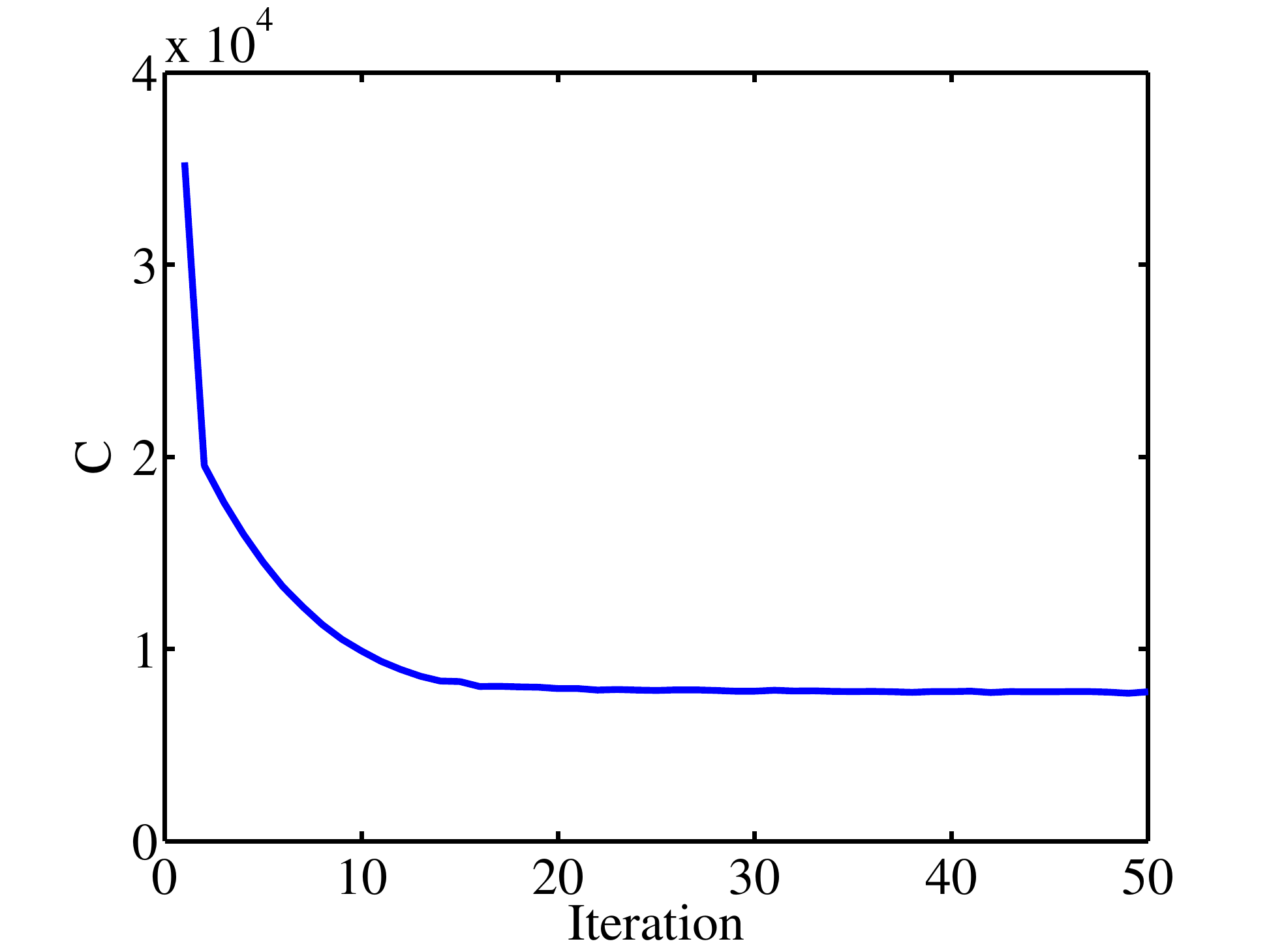}
\caption[Cost changes during the optimization process.]{Cost changes during the optimization process.  The experiment is conducted on NUS-WIDE data set with $d=64$ and $\alpha=0.1$.  From the figure we can see that the optimization converges quickly and there are almost no changes after the first 20 iterations.}
\label{fig:costauto}
\end{figure}

There are two parameters that need to be chosen for {\it Auto-Jacobin}: the number of iterations of the line search optimization process and the weight parameter $\alpha$. We have analyzed the costs during optimization and notice that it converges quickly. As a case in point, Fig.~\ref{fig:costauto} shows the cost after each iteration. The experiment is conducted on NUS-WIDE data set. The task is to learn $64$ bits for encoding the base data set, and the weight parameter is set to be $0.1$. From the figure, we can see that the objective function converges quickly and it makes tiny changes after the first $20$ iterations. Another observation from the cost-iteration plot is that the cost decreases monotonically despite lack of guarantees that this would occur since the input data points keep changing in each iteration. Thus, we can either set it to be a fixed value as done in our experiments or terminate when the cost is below a threshold or set it to stop when the change in cost over recent iterations drops below some threshold. In our experiments, the number of training data points is $10,000$, the batch size is $1,000$, and the training data points are randomly shuffled in each epoch. Thus, each pass of the training data set takes $m=10$ iterations. We set the total number of iterations to be $50$ in accordance with other hashing methods. Therefore, the optimization cycles through the training data set $I_{max}=5$ times in total.

\begin{figure}[htbp!]
\centering
\includegraphics[width=8cm]{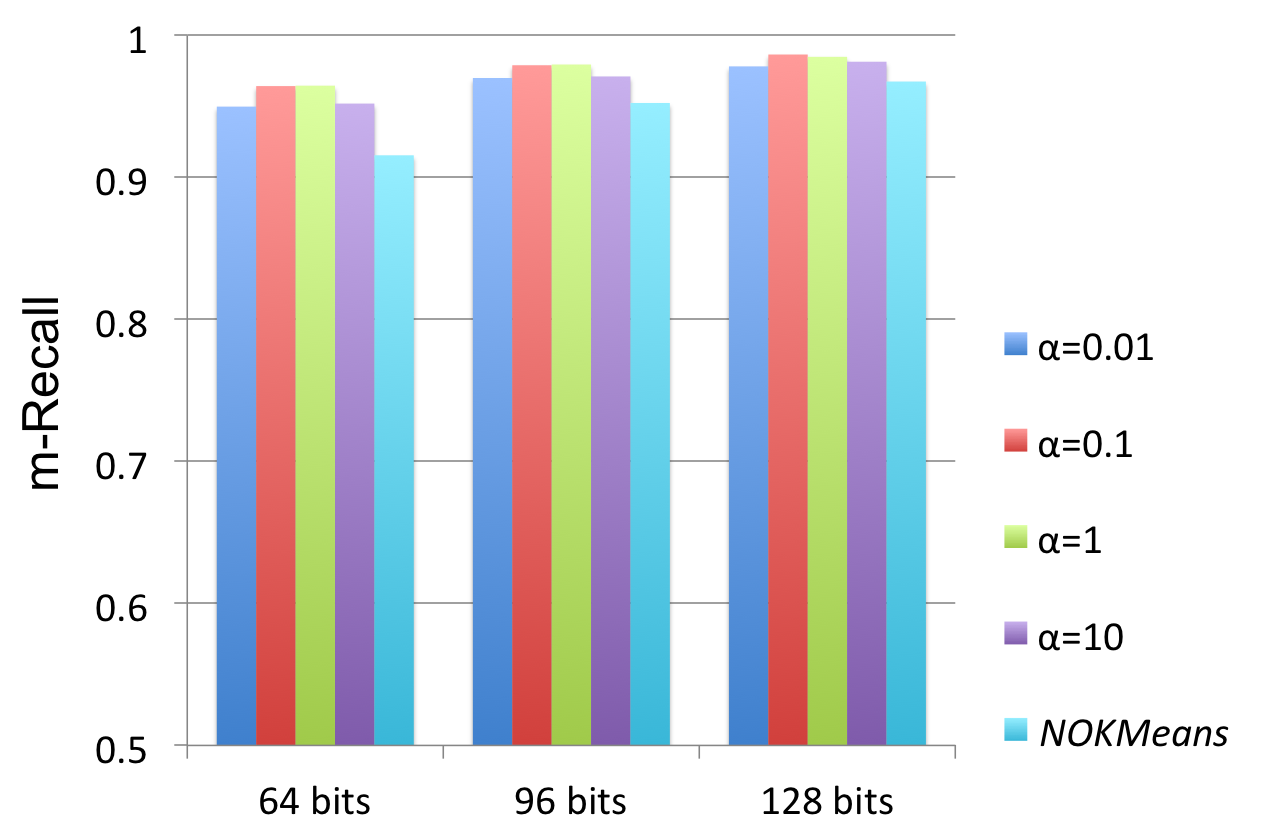}
\caption[The m-Recall performance when the weight parameter ranges from $0.1$ to $10$.]{The m-Recall performance when the weight parameter ranges from $0.1$ to $10$. The computational task is to find the nearest neighbor point for each query in the base data set of NUS-WIDE. The performance is compared when each data point is encoded into $64, 96$ and $128$ bits respectively. }
\label{fig:m-Recall}
\end{figure}

For the weight parameter $\alpha$, we empirically test its value at $0.01, 0.1, 1, 10$. Fig.~\ref{fig:m-Recall} shows the m-Recall when the data points are encoded with $64, 96$ and $128$ bits respectively for the nearest point search for each query in the NUS-WIDE data set. For visual comparison, we have included the corresponding m-Recall performance of {\it NOKMeans} which is a competing hashing method as shown in the experiments later. From the chart in Fig.~\ref{fig:m-Recall}, we can see that the proposed approach has stable performance  when the weight parameter $\alpha$ ranges from $0.01$ to $10$, and its overall performance is always better than {\it NOKMeans} on the NUS-WIDE data set. In the following experiments, we fix the weight parameter $\alpha$ to be $0.1$.

\subsection{Performance with different auto-encoder models}
In order to see the impact of the first order constraint for {\it Auto-JacoBin}, we report the comparative results for AutoBin - an optimization without the Jacobian component. The cost function of AutoBin is optimized with two different methods. One, referred to as {\it AutoBin1},  uses the same optimization method as in {\it Auto-JacoBin}. The other, referred to as {\it AutoBin2}, is using the whole training data set in each iteration.

The proposed auto-encoder is motivated by {\it DAEs} and {\it CAEs} which are designed to have a noise removing effect. It is also interesting to see the performance of the corresponding hashing methods when the auto-encoder component is replaced with {\it DAEs} and {\it CAEs} respectively. We refer these two new hashing methods as denoising auto-encoder binary hashing ({\it DAutoBin}) and contractive auto-encoder binary hashing ({\it CAutoBin}). For {\it DAutoBin}, the noise is injected by randomly setting a fixed percentage of entries of original data points as 0 as it is used in \cite{vincent2010stacked}. Take the 960D GIST feature as an example, suppose we fix the percentage by threshold $t$, we generate a random number $r_i$ ($i=1, 2, \cdots, 960$) between $0$ and $1$ for each entry of $x=(x_1, x_2, \cdots, x_{960})' \in \mathcal{R}^{960}$,  and set 
\begin{equation}
x_i=\left\{
\begin{array}{cc}
x_i,& r_i > t\\
0,& r_i \leq t
\end{array}
\right.
\end{equation}
We empirically test the parameter $t$ from a set $\{0.01, 0.05, 0.1, 0.2\}$ and the weight parameter $\alpha$ from a set $\{0.01, 0.1, 1, 10\}$. 
For {\it CAutoBin}, both the parameter $\lambda$ for the Jacobian norm and the weight parameter $\alpha$ are tuned from the set $\{0.01, 0.1, 1, 10\}$. Finally, the optimization for {\it DAutoBin} and {\it CAutoBin} is similar to Algorithm~\ref{alg:AutoHashing}.  The main difference is the gradient calculations which can be derived in the same way as for {\it Auto-JacoBin}. 

\begin{figure}[htb]
\centering
\includegraphics[width=8.0cm]{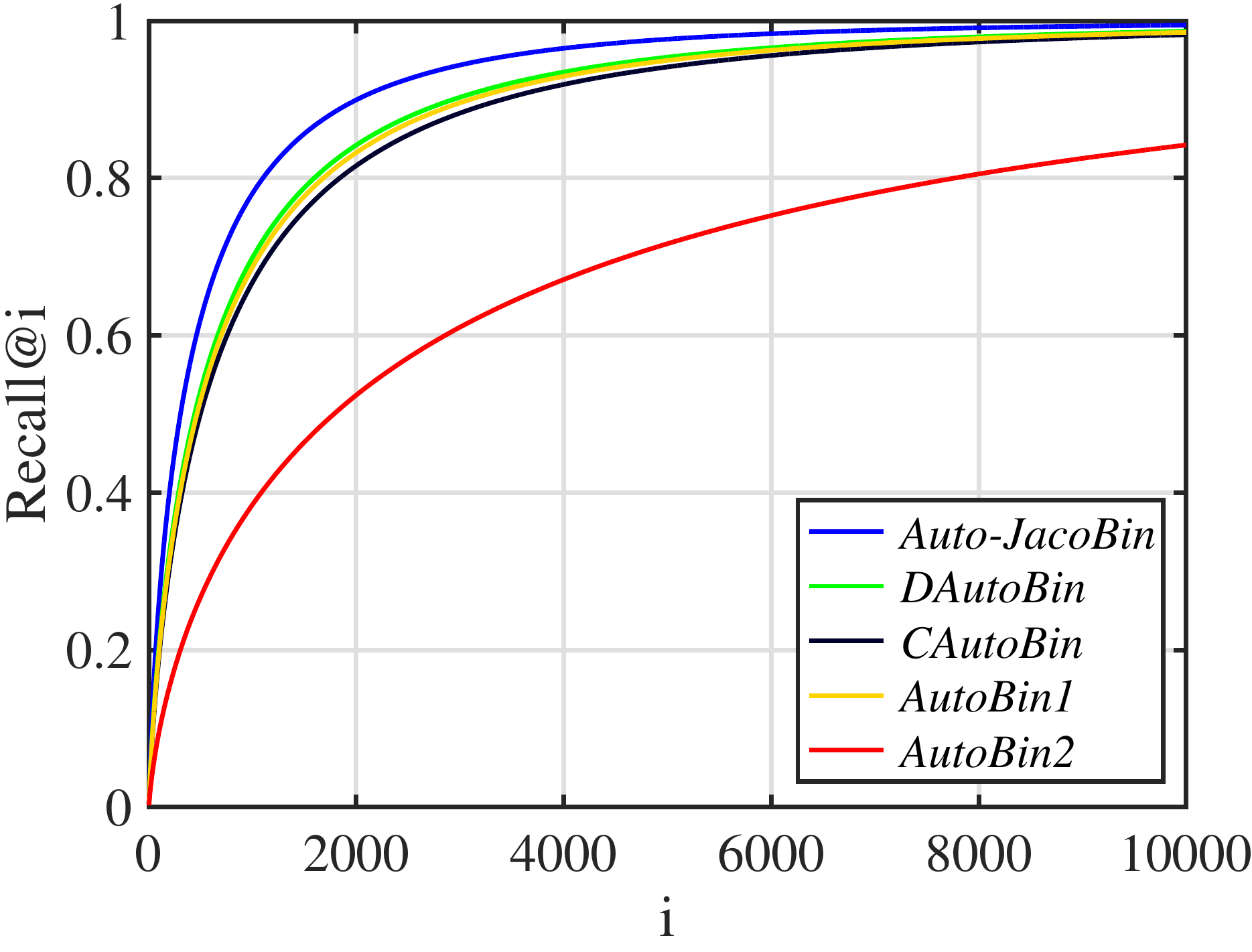}
\caption[Retrieval performance on NUS-WIDE data set by different auto-encoder models used in the optimization objective.]{{Retrieval performance on NUS-WIDE data set by different auto-encoder models used in the optimization objective. Each data point is encoded with 128 bits, and the retrieval task is to find the 100 nearest neighbor points for each query. }}
\label{fig:NUSWIDE_differentAuto}
\end{figure}

Fig.~\ref{fig:NUSWIDE_differentAuto} shows the performance of the hashing methods based on different auto-encoder models on NUS-WIDE. Each data point is encoded with $128$ bits, $k$ is set to $100$, and the $\text{Recall@}i$ performance is reported  when $i$ ranges from $1$ to $10,000$. 
From the figure, we can see that the $\text{Recall@}i$ of {\it Auto-JacoBin} is consistently better than the performance of other hashing methods. {\it DAutoBin}, {\it CAutoBin} and {\it AutoBin1} have comparable results, and {\it AutoBin2} gives the lowest $\text{Recall@}i$.  Although the objective function of {\it AutoBin1} and {\it AutoBin2} is the same, their performance is quite different, which can be attributed to mini-batch stochastic gradient descent being less likely to get stuck in local optima.

\begin{figure*}[htbp!]
\centering
\includegraphics[width=16.7cm,height= 17.5cm]{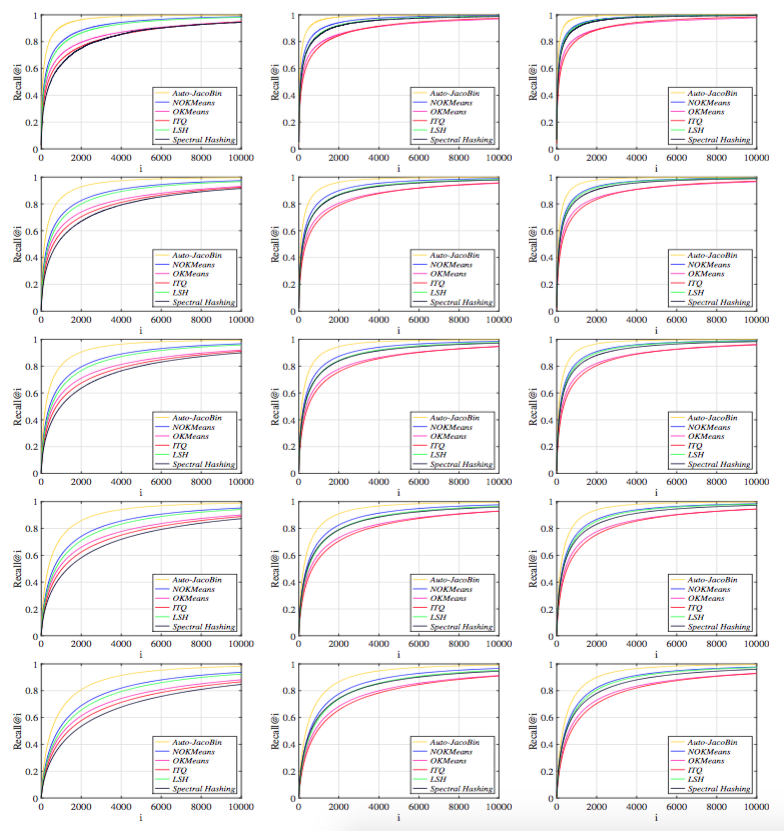}
\caption[Retrieval performance on NUS-WIDE data set.]{Retrieval performance on NUS-WIDE data set. The rows from top to bottom show retrieval performance for $k=\{1, 5, 10, 50, 100\}$ (nearest neighbours). The columns from left to right show the performance with $64, 96, 128$ encoding bits respectively.}
\label{fig:NUSWIDE_recall}
\end{figure*}

From our experiments, we note that, {\it DAutoBin} is slightly better than {\it AutoBin1}, and {\it AutoBin1} is slightly better than {\it CAutoBin}. The $\alpha$ in {\it CAutoBin} is set to be $0.01$, we found that the performance of {\it CAutoBin} decreases when $\alpha$ increases in our experiments. The possible reason is that the denoising effect in {\it CAutoBin} is achieved implicitly by balancing two components. Due to the contractive component, the recovery of the original data points is not as good as using auto-encoder directly. Thus the performance of {\it CAutoBin} is decreased slightly. For {\it DAutoBin}, the noise is injected manually and the neural network is trained to recover the data point. One of the problems for this approach is that it is unclear why the original data is the best recovery of the noisy data. Here is an extreme case, suppose $x_1$ and $x_2$ are two close but different data points, thus they might have the same noisy data point $\overline x$. According to the definition, when $\overline x$ is an input data point,  the output of {\it DAutoBin} is assumed to be $x_1$ and $x_2$ at the same time. This contradicts the fact that $x_1$ and $x_2$ are two distinct points.  On the other hand, {\it Auto-JacoBin} does not have this kind of contradictory behavior since it estimates the best projection of the noisy data point. We believe this is one of the reasons that {\it Auto-JacoBin} has a better performance than {\it DAutoBin}.

\subsection{Results on Benchmark data sets}
\begin{figure*}[htb]
\centering
\includegraphics[width=16.7cm,height= 17.5cm]{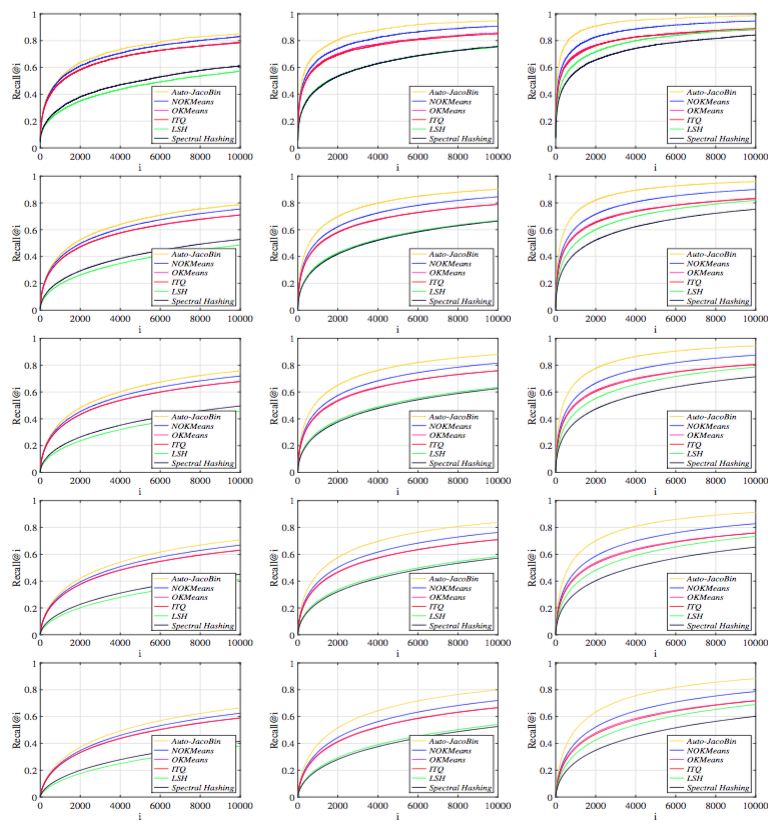}
\caption[Retrieval performance on GIST1M data set.]{Retrieval performance on GIST1M data set.
The rows from top to bottom show retrieval performance for $k=\{1, 5, 10, 50, 100\}$ (nearest neighbours). The columns from left to right show the performance with $64, 128, 256$ encoding bits respectively.}
\label{fig:GIST_recall}
\end{figure*}

The computational task is to find the $k$ nearest neighbor points in the base data set for each query.
Here we evaluate scenarios where $k\in \{1, 5, 10, 50, 100\}$. The $\text{Recall@}i$ performance is reported  when $i$ ranges from $1$ to $10,000$. Therefore, only a part of the base data set is retrieved ($1\%$, $0.1\%$ and $0.1\%$ for the NUS-WIDE, GIST1M and SIFT1M respectively). For both NUS-WIDE and SIFT1M, we report the $\text{Recall@}i$ performance when the number of bits, which are used to encode the data points, is up to its feature's dimension, i.e., they are tested with $64$, $96$ and $128$ bit encodings. For the GIST1M data set, more bits are used to encode the data points because the underlying features have more dimensions. In our experiments, GIST1M is tested with $64, 128$ and $256$ bit encodings. The results for NUS-WIDE are shown in Fig.~\ref{fig:NUSWIDE_recall} and the results for GIST1M is in Fig.~\ref{fig:GIST_recall}. The $\text{Recall@}i$ performance of different hashing approaches for SIFT1M are much closer and more difficult to see in a graph, and therefore we present the m-Recall in a table. Table~\ref{table:sift1M} shows the  m-Recall performance of different hashing methods on the SIFT1M data set. Each row corresponds to one specific setting, where $(i,j)$ means the retrieval task is to find the $i$ nearest neighbor points for each query and $j$ bits are used for encoding the data set. 

\begin{table*}[htb]
\begin{center}
\begin{tabular}{|c|ccccc|c|}
\hline
~~~~Algorithms~~~~&  {\it ~~~~~ITQ~~~~~} & {\it ~~~~LSH~~~~} & {\it Spectral Hashing} & {\it ~~OKMeans~~} & {\it ~~NOKMeans~~} & {\it ~Auto-JacoBin~} \\
\hline
  (1, 64) &0.879 	 &0.825 	 &0.908 	 &0.933 	 &0.930 	 &{\bf 0.935} \\  
  (1, 96) &0.940 	 &0.912 	 &0.944 	 &0.967 	 &0.967 	 &{\bf 0.971} \\  
  (1, 128) &0.965 	 &0.952 	 &0.962 	 &0.979 	 &0.982 	 &{\bf 0.983} \\  
 (5, 64) &0.842 	 &0.772 	 &0.865 	 &0.903 	 &0.893 	 &{\bf 0.905} \\  
 (5, 96) &0.915 	 &0.872 	 &0.910 	 &0.947 	 &0.943 	 &{\bf 0.952} \\  
 (5, 128) &0.946 	 &0.926 	 &0.936 	 &0.965 	 &0.966 	 &{\bf 0.971} \\  
 (10, 64) &0.820 	 &0.740 	 &0.837 	 &0.882 	 &0.867 	 &{\bf 0.883} \\  
 (10, 96) &0.897 	 &0.847 	 &0.887 	 &0.931 	 &0.924 	 &{\bf 0.937} \\  
 (10, 128) &0.933 	 &0.907 	 &0.916 	 &0.953 	 &0.953 	 &{\bf 0.960} \\  
 (50, 64) &0.752 	 &0.650 	 &0.749 	 &{\bf 0.814} 	 &0.779 	 &0.811 \\  
 (50, 96) &0.839 	 &0.767 	 &0.806 	 &0.876 	 &0.853 	 &{\bf 0.881} \\  
 (50, 128) &0.884 	 &0.842 	 &0.844 	 &0.908 	 &0.898 	 &{\bf 0.915} \\  
 (100, 64) &0.714 	 &0.604 	 &0.699 	 &{\bf 0.773} 	 &0.728 	 &0.767 \\  
 (100, 96) &0.803 	 &0.722 	 &0.757 	 &0.840 	 &0.807 	 &{\bf 0.843} \\  
 (100, 128) &0.852 	 &0.802 	 &0.797 	 &0.876 	 &0.859 	 &{\bf 0.883} \\ \hline
\end{tabular}
\end{center}
\caption[Retrieval performance (m-Recall) of different hashing methods on SIFT1M.]{Retrieval performance (m-Recall) of different hashing methods on SIFT1M. Each row corresponds to one specific setting, where $(i,j)$ means $j$ bits are used for encoding the data set and the retrieval task is to the find $i$ nearest neighbor points for each query.  For these tests, $K=10,000$.}\label{table:sift1M}
\end{table*}

From Fig.~\ref{fig:NUSWIDE_recall}  and Fig.~\ref{fig:GIST_recall}, it is clear that the proposed method performs much better than previous techniques. Other observations are that (unsurprisingly) using more bits results in better performance for all methods, and that the performance decreases as $k$ increases. Somewhat surprisingly, the data independent {\it LSH} algorithm performs comparably with many of the data-dependent algorithms, especially as the number of bits increases.
One of the main reasons is that  when modeling the data distribution, the data dependent hashing algorithms are done by some kind of relaxation such as obtaining the binary code from some threshold or encoding the data point into its nearest binary code. It is inevitable that the information contained in the original data set is lost to some degree when a limited number of bits are used to encode each data point during this process. While for {\it LSH}, the obtained binary codes can preserve the similarity between the data points even though the projection is randomly generated. 
Another interesting point is that the comparative performance of different hashing algorithms depends on the data set. For NUS-WIDE, when few bits are used to encode the data points, {\it Spectral Hashing} has the lowest score. When the number of bits is increased, the performance of {\it Spectral Hashing} is comparable to {\it LSH} and {\it NOKMeans}, and has much better performance than {\it ITQ} and {\it OKMeans}. Both {\it ITQ} and {\it OKMeans} assume that their projection matrix is orthogonal. This constraint limits the separating capacity of different bits. That is why when more bits are used for encoding of the data points, the performance does not increase as much as other algorithms. For the proposed hashing method and {\it Spectral Hashing} and {\it LSH}, the projection matrices do not have this kind of constraint. Thus the potential partition capability of these algorithms is better and they show more consistent performance.

Overall, the benefits of the proposed method are clear.
Even with a limited bit budget, the proposed hashing method has excellent $\text{Recall@}i$ performance. For instance, when 64 bits are used to encode each data point on the NUS-WIDE data set, $\text{Recall@}1000$ is $0.92$, whilst the previous best is {\it NOKMeans} at 0.81. Also, to ensure good recall, when using 256 bits on the GIST1M data set, the proposed method requires $K=2,000$ to achieve a 0.9 $\text{Recall@}i$. Compared with $K=6,000+$ for the other methods, ours allows for improved computational performance in practice. 

For the local feature data set, the proposed method has comparable performance to state-of-the-art results ({\it NOKMeans}, {\it OKMeans} and {\it ITQ}). For better visualization, we summarize some of the results based on the $\mbox{m-Recall}$ indicator. Table~\ref{table:sift1M} shows the  $\mbox{m-Recall}$ when the data points are encoded with different bits and for different retrieving tasks.
From this table, we can see that the proposed method has comparable performance with other state-of-the-art algorithms. Its performance becomes better when more bits are used to encode the data points. 
Another observation we can make is that, despite its good performance on NUS-WIDE and GIST1M, {\it LSH} is consistently the worst, or nearly the worst, performer for this data set. 

\section{Summary}
In this work, we have proposed a novel hashing algorithm which adopts the auto-encoder model to produce binary codes that are geometrically consistent with the data points in the original space.  The model we used leverages the merits of state-of-the-art models for auto-encoders. We have introduced a new objective function that makes use of the first order properties of geometric conservation (the Jacobian matrix) and has good noise-reduction properties. We proved that the Jacobian of the defined function can be used to approximate the tangent space of the manifold of the training data set, and show how the Jacobian can be expressed analytically. 

The experiments are conducted on three large scale feature data sets. The performance of the proposed method is compared with several state-of-the-art hashing algorithms. It has the best performance for global image features and comparable performance for the SIFT feature data sets. 

Currently, we are interested in binary codes for large-scale high-dimensional data sets. Thus, we adapt the proposed auto-encoder model by adding constraints in the hidden layer. In the future, we plan to investigate the applications of the new auto-encoder model in other  feature learning fields including dimensionality reduction and deep learning.

\onecolumn
\section*{Appendix I}
%\begin{theorem}\label{mcase}
%Suppose $\mathcal{M}$ is a $d$ dimensional compact smooth submanifold in $\mathcal{R}^D$, $f$ is a function defined as $f(x)=\argmin_{m\in \mathcal{M}}||x-m||_2^2$,  $\forall x \in \mathcal{R}^D$ . For each $m$ in $\mathcal{M}$, let $T_m\in \mathcal{R}^{D\times d}$ be the local normal basis of the tangent space to $\mathcal{M}$ at $m$. Then, the Jacobian matrix of $f$ at $m$ is $T_mT_m'$.
%\end{theorem}
\noindent\textbf{Proof of Theorem 1.}~~
To prove $J_m=T_mT_m'$, we have to show that
 \begin{equation}\label{eq:f1}
\lim_{t \rightarrow 0}\frac{||f(m+t)-f(m)-T_mT_m't||}{||t||}=0,
 \end{equation}
 where $m\in \mathcal{M} \subset \mathcal{R}^D$ and $t \in \mathcal{R}^D$.

\begin{figure}[ht!]
\centering
\includegraphics[width=8cm]{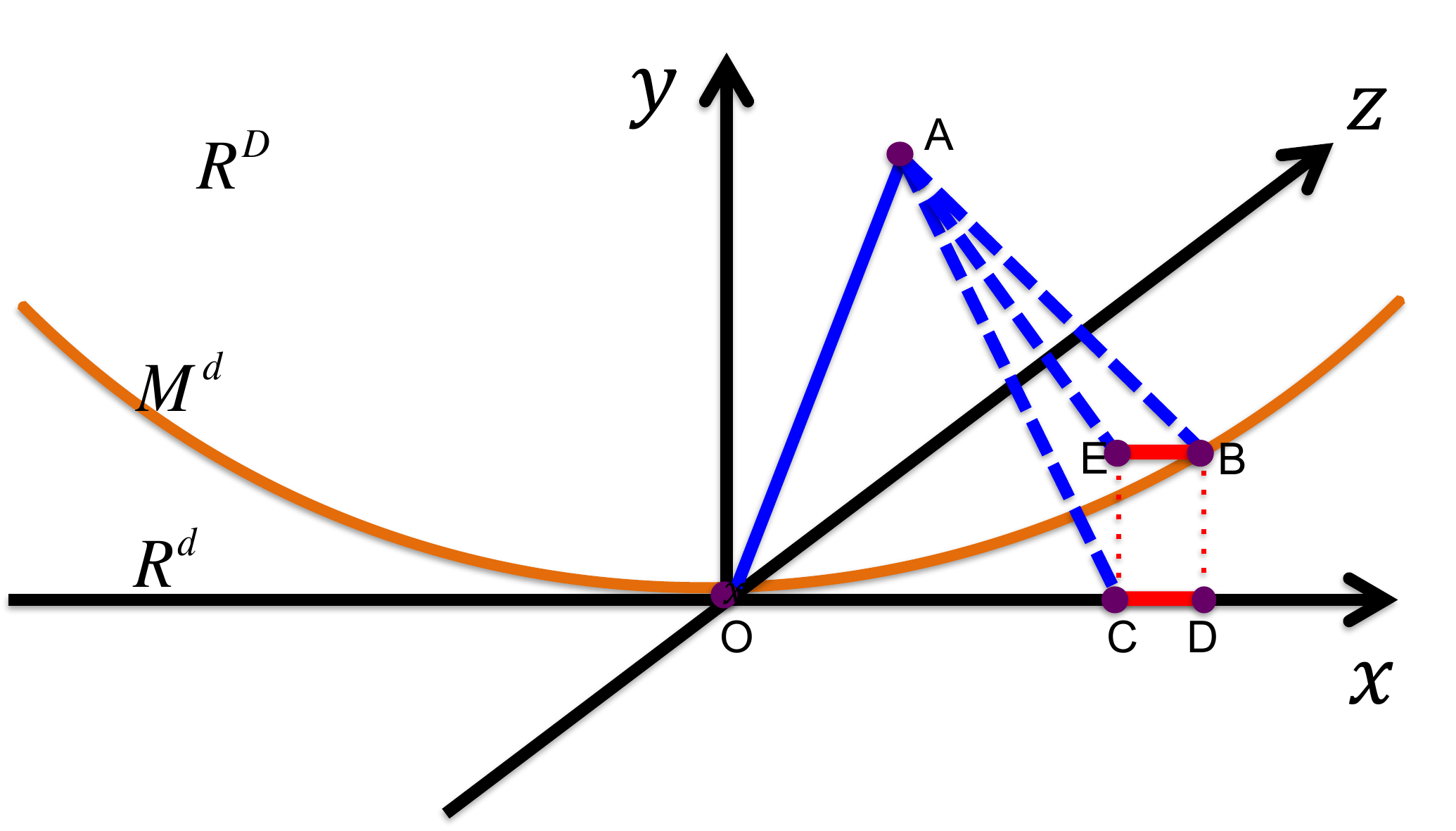}
\caption[Visualization of the positions in Euclidean space.]{Visualization of the positions in Euclidean space. For better understanding, the yellow curve can be imaginged as a $d$ dimensional manifold, and the $x$-axis is viewed as a tangent plane at position $O$.
}
\label{fig:stable_auto_encoder}
\end{figure}

Since $m \in \mathcal{M}$, from the definition of $f$ we can see that $f(m)=m$. Thus $f(m)+T_mT_m't$ is the projection point of $m+t$ to the tangent space at $m$. Denote the positions of the points $m$, $m+t$, and $f(m+t)$ as $O$, $A$, and $B$ respectively.  Let $C$ and $D$ be the projections of  $A$ and $B$ to the tangent space respectively, as shown in Fig.~\ref{fig:stable_auto_encoder}. Denote $UV$ as the line segment between $U$ and $V$, then $\overrightarrow{UV}$ and $||UV||$ are the corresponding vector and the length of the segment.
Thus, Equation \eqref{eq:f1} is equivalently represented as:
 \begin{equation}\label{eq:f2}
\lim_{t \rightarrow 0}\frac{||CB||}{||OA||}=0.
 \end{equation}

In the following, we show that
 \begin{equation}\label{eq:f3}
\lim_{t \rightarrow 0}\frac{||CD||}{||OA||}=0, 
 \end{equation}
and
 \begin{equation}\label{eq:f4}
\lim_{t \rightarrow 0}\frac{||BD||}{||OA||}=0.
 \end{equation}

Since $CB$ is a side of the  triangle $CDB$, we have $||CB||\leq ||CD||+||DB||$. Thus, the results from Equation \eqref{eq:f3} and Equation \eqref{eq:f4} ensure that
Equation \eqref{eq:f2} is true.

To prove Equation \eqref{eq:f3}, we first locate the point $E$, such that $\overrightarrow{EB}=\overrightarrow{CD}$. Note that $D$ is the projection point of $B$ to the tangent space, and $\overrightarrow{CD}$ is a vector in the tangent space, thus we have $\overrightarrow{BD}\perp \overrightarrow{CD}$. It is easy to show that the quadrilateral EBDC is a rectangle. Thus, we can conclude that $\overrightarrow{EB}\perp \overrightarrow{EC}$. On the other hand, since $C$ is the projection point of $A$ in the tangent space, we have $\overrightarrow{AC}\perp \overrightarrow{CD}$. Therefore, we have $\overrightarrow{AC}\perp \overrightarrow{EB}$ since $\overrightarrow{EB}=\overrightarrow{CD}$. Considering the plane $ACE$, we have
$\overrightarrow{AC}\perp \overrightarrow{EB}$ and $\overrightarrow{EC}\perp \overrightarrow{EB}$, therefore $\overrightarrow{EB} \perp$ the plane $ACE$ and $\overrightarrow{EB}\perp \overrightarrow{AE}$.

Consider $\angle ABE$. $B$ is the closest point on $\mathcal{M}$ to $A$, so
\begin{equation}
||OB||\leq||OA||+||AB||\leq2||OA||,
\end{equation}
and $||OB|| \rightarrow 0$ when $t \rightarrow 0$. Therefore, the tangent space at $B$ tends to the tangent space at $O$ when $t \rightarrow 0$ since $\mathcal{M}$ is a smooth submanifold in $\mathcal{R}^D$. From the definition of $f$, we can conclude that $AB$ is perpendicular to $T_B$. Thus, $\angle ABE \rightarrow \frac{\pi}{2}$, when $t \rightarrow 0$. Since $\overrightarrow{EB}\perp \overrightarrow{AE}$, we have 
\begin{equation}||CD||=||EB||=||AB|| \cos \angle ABE\leq ||OA||\cos \angle ABE. \end{equation} 
The final inequality is true due to the definition of the function $f$.
Thus we have \begin{equation}0\leq \lim_{t \rightarrow 0}\frac{||CD||}{||OA||}\leq\lim_{t \rightarrow 0} \cos \angle ABE=0,\end{equation} and so Equation \eqref{eq:f3} is proved.

To prove Equation \eqref{eq:f4}, we first show that 
\begin{equation}\label{eq:f11}
\lim_{t \rightarrow 0}\frac{||BD||}{||OD||}=0
\end{equation}
Since $\mathcal{M}$ is smooth, the first order of the manifold structure is continuous. Thus we have $\angle BOD\rightarrow 0$ when $t\rightarrow 0$. 
Since $\frac{||BD||}{||OD||}$ is exactly the tangent of $\angle BOD$, we have the conclusion that $\lim_{t \rightarrow 0}\frac{||BD||}{||OD||}=0$.

Since $\lim_{t \rightarrow 0}\frac{||CD||}{||OA||}=0$ and $||OC||\leq ||OA||$, $\forall \epsilon>0$, $\exists \delta$, such that when $||t||<\delta$, we have $0\leq\frac{||OC||+||CD||}{||OA||}<1+\epsilon$. Therefore, when $||t||<\delta$, we have
\begin{align}
0\leq\lim_{t \rightarrow 0}\frac{||BD||}{||OA||} \leq &(1+\epsilon)\lim_{t \rightarrow 0}\frac{||BD||}{||OC||+||CD||}\\
\leq& (1+\epsilon)\lim_{t \rightarrow 0}\frac{||BD||}{||OD||}=0,\notag
\end{align}
and so Equation \eqref{eq:f4} is proved. ~~~~~~~~~~~~~~~~~~~~~~~~~~~~~~~~~~~~~~~~~~$\square$
\vspace{0.3cm}

\onecolumn
\section*{Appendix II}
\label{appendix}
\subsection{Objective function}
The objective function for {\it Auto-JacoBin} is: 
 \begin{equation*}
 \scriptsize
 \begin{split}
\mathcal{C}_3(W_1,W_2,b_1,b_2)=&\sum_{i=1}^{N}(||x_i-z_i||_F^2+||J_i-T_iT_i'||_F^2) +\alpha||YY'-NI||_1^\epsilon.
\end{split}
 \end{equation*}

In the following gradients, $\odot$ is the operator of point-wise product between two matrices.  To unveil the steps of the gradients calculation, we denote $W_2^{(i)}$ as the $i^{\mbox{\scriptsize th}}$ row vector of $W_2$, $1_i^{n}$ as the zero column vector with the $i^{\mbox{\scriptsize th}}$ position equal to $1$, $O^n$ as an $n$-dimensional one vector, and $\delta_{n,m}^{W_1}$ as the zero matrix with size $W_1$ with its $(n,m)^{\mbox{\scriptsize th}}$ position $1$. We also use $\otimes$, $\ominus$ and $sum(\cdot)$ operators, which function as follows. The $W_1\otimes W_2$ is a $D\times D \times d$ matrix such that its $(i, j, k)^{\mbox{\scriptsize th}}$ element is $W_1^{(ki)}W_2^{(jk)}$. For a $D\times D$ matrix $U$ and a $D\times D\times d$  matrix $V$, $U \ominus V$ is a  $D\times D\times d$ matrix such that its $(i, j, k)^{\mbox{\scriptsize th}}$ element is $U^{(ij)}V^{(ijk)}$. The operation of both   $\otimes$ and $\ominus$ can be implemented in Matlab by the $repmat$ and $dot$ functions efficiently. For a 2-D or 3-D matrix $Q$, the $sum(Q)$ produces respectively a 1-D or 2-D matrix by adding the row entries similarly to the $sum $ function in Matlab.

\subsection{Auto-encoder constraint}
The gradients can be calculated as:
 \begin{equation*}
 \scriptsize
 \begin{split}\frac{\partial ||z-x||_F^2}{\partial W_1}=2\left(\left(W_2'\left(\left(z-x\right)\odot \left(1-z^2\right)\right)\right)\odot\left(1-y^2\right)\right)x'~~~~~~~~~~~~~~~~~~~~~~~~~~~~~~~~~~~~~~~~~~~~~~~~~~~~~~~~~
 \end{split}
 \end{equation*}

 \begin{equation*}
 \scriptsize
 \begin{split}\frac{\partial ||z-x||_F^2}{\partial b_1}=2\left(\left(W_2'\left(\left(z-x\right)\odot \left(1-z^2\right)\right)\right)\odot\left(1-y^2\right)\right)~~~~~~~~~~~~~~~~~~~~~~~~~~~~~~~~~~~~~~~~~~~~~~~~~~~~~~~~~~~~
\end{split}
 \end{equation*}

 \begin{equation*}
 \scriptsize
 \begin{split}\frac{\partial ||z-x||_F^2}{\partial W_2}=2\left(\left(z-x\right)\odot \left(1-z^2\right)\right)y'~~~~~~~~~~~~~~~~~~~~~~~~~~~~~~~~~~~~~~~~~~~~~~~~~~~~~~~~~~~~~~~~~~~~~~~~~~~~~~~~~~~~~~
\end{split}
 \end{equation*}

 \begin{equation*}
 \scriptsize
 \begin{split}\frac{\partial ||z-x||_F^2}{\partial b_2}=2\left(\left(z-x\right)\odot \left(1-z^2\right)\right)~~~~~~~~~~~~~~~~~~~~~~~~~~~~~~~~~~~~~~~~~~~~~~~~~~~~~~~~~~~~~~~~~~~~~~~~~~~~~~~~~~~~~~~
\end{split}
 \end{equation*}

\subsection{Binary constraint}
 \begin{equation*}
 \scriptsize
 \begin{split}\frac{\partial ||YY'-NI||_1^\epsilon}{\partial W_1}=2\left(\left(\left(YY'-NI\right)\left(\left(YY'-NI\right)^2+\epsilon\right)^{-\frac{1}{2}}Y\right)\odot\left(1-Y^2\right)\right)X'
~~~~~~~~~~~~~~~~~~~~~~~~~~~~~
\end{split}
 \end{equation*}
 \begin{equation*}
 \scriptsize
 \begin{split}\frac{\partial ||YY'-NI||_1^\epsilon}{\partial b_1}=2\left(\left(\left(YY'-NI\right)\left(\left(YY'-NI\right)^2+\epsilon\right)^{-\frac{1}{2}}Y\right)\odot\left(1-Y^2\right)\right)~~~~~~~~~~~~~~~~~~~~~~~~~~~~~~~~~
 \end{split}
 \end{equation*}

\subsection{First order constraint}
To simplify the notation, we denote $A=TT'$. Since the constraint can be decomposed as $||J-A||_F^2=||J||_F^2-2Tr(JA')+const$, the gradients can be decomposed into two components:

\subsubsection{Gradients for $||J||_F^2$}
Since $z$ is a vector, we can calculate its gradient according to each component $z^{(i)}$:
 \begin{equation*}
 \scriptsize
 \begin{split}\frac{\partial(z^{(i)})}{\partial x}=\frac{\partial(\tanh(W_2^{(i)}\tanh(W_1x+b_1)+b_2^{(i)}))}{\partial x}=W_1'\left(\left(\left(W_2^{(i)}\right)'\left(1-(z^{(i)})^2\right)\right)\odot\left(1-y^2\right)\right)
\end{split}
 \end{equation*}

The Jacobian function can be expressed as
 \begin{equation*}
\scriptsize
 \begin{split}J=W_1'\left( W_2'\odot\left(1-y^2\right)\left(1-z^2\right)'\right)\end{split}
 \end{equation*}

and  \begin{equation*}
\scriptsize
 \begin{split}||J||_F^2=||\frac{\partial(z)}{\partial x}||_{F}^2=\sum_{i=1}^{D}\left(1-(z^{(i)})^2\right)^2\sum_{m=1}^{D}
\left(\sum_{n=1}^{d}W_1^{(n,m)}W_2^{(i,n)}\left(1-(y^{(n)})^2\right)\right)^2\end{split}
 \end{equation*}

Thus
\begin{equation*}
\scriptsize
 \begin{split}
 \frac{\partial ||J||_F^2}{\partial W_1}&=\sum_{i=1}^{D}\frac{\partial\left(1-(z^{(i)})^2\right)^2}{\partial W_1}\sum_{m=1}^{D}
\left(\sum_{n=1}^{d}W_1^{(n,m)}W_2^{(i,n)}\left(1-(y^{(n)})^2\right)\right)^2\\
&~~~~~+\sum_{i=1}^{D}\left(1-(z^{(i)})^2\right)^2\sum_{m=1}^{D}\frac{\partial
\left(\sum_{n=1}^{d}W_1^{(n,m)}W_2^{(i,n)}\left(1-(y^{(n)})^2\right)\right)^2}{\partial W_1}\\
&=\sum_{i=1}^{D}\left(\left(\left(W_2^{(i)}\right)'2\left(1-(z^{(i)})^2\right)^2\left(-2z^{(i)}\right)\right)\odot\left(1-y^2\right)\right)x'\sum_{m=1}^{D}
\left(\sum_{n=1}^{d}W_1^{(n,m)}W_2^{(i,n)}\left(1-(y^{(n)})^2\right)\right)^2\\
&~~~~~+ \sum_{i=1}^{D}\left(1-(z^{(i)})^2\right)^2\sum_{m=1}^{D}2
\left(\sum_{n=1}^{d}W_1^{(n,m)}W_2^{(i,n)}\left(1-(y^{(n)})^2\right)\right)\left(\sum_{n=1}^{d}\delta_{n,m}^{W_1}W_2^{(i,n)}\left(1-(y^{(n)})^2\right)\right)\\
&~~~~~+ \sum_{i=1}^{D}\left(1-(z^{(i)})^2\right)^2\sum_{m=1}^{D}2
\left(\sum_{n=1}^{d}W_1^{(n,m)}W_2^{(i,n)}\left(1-(y^{(n)})^2\right)\right)\\
&~~~~~\times\left(\sum_{n=1}^{d}1_n^{d}\left(W_1^{(n,m)}W_2^{(i,n)}\left(-2y^{(n)}\right)\left(1-(y^{(n)})^2\right)\right)x'\right)\\
&=-4\left(W_2'\odot \left(O^{d}\left(\left(1-z^2\right)^2\left(z\right)\right)'\right)\odot \left(\left(1-y^2\right)\left(O^{D}\right)'\right)\odot \left(O^{d}sum\left(\left(W_1'\left(W_2'\odot\left(1-y^2\right)(O^{D})'\right)\right)^2\right)\right)\right)O^{D}x'\\
&~~~~~+2\left(W_2'\left(W_1'\left(W_2'\odot \left(O^{d}\left(\left(1-z^2\right)^2\right)'\right)\odot \left(\left(1-y^2\right)(O^{D})'\right)\right)\right)'\right)\odot\left(\left(1-y^2\right)\left(O^{D}\right)'\right)\\
&~~~~~-4\left(sum\left(sum\left(\left(\left(O^{D}\left(\left(1-z^2\right)^2\right)'\right)\odot\left(W_1'\left(W_2'\odot\left(1-y^2\right)\left(O^{D}\right)'\right)\right)\right)\ominus \left(W_1\otimes W_2\right)\right)\right)\right)'\\
&~~~~~\odot\left(y-y^3\right)x'\end{split}
 \end{equation*}

\begin{equation*}
\scriptsize
\begin{split}
 \frac{\partial ||J||_F^2}{\partial b_1}&=\sum_{i=1}^{D}\frac{\partial\left(1-(z^{(i)})^2\right)^2}{\partial b_1}\sum_{m=1}^{D}
\left(\sum_{n=1}^{d}W_1^{(n,m)}W_2^{(i,n)}\left(1-(y^{(n)})^2\right)\right)^2\\
&~~~~~+\sum_{i=1}^{D}\left(1-(z^{(i)})^2\right)^2\sum_{m=1}^{D}\frac{\partial
\left(\sum_{n=1}^{d}W_1^{(n,m)}W_2^{(i,n)}\left(1-(y^{(n)})^2\right)\right)^2}{\partial b_1}\\
&=\sum_{i=1}^{D}\left(\left(\left(W_2^{(i)}\right)'2\left(1-(z^{(i)})^2\right)^2\left(-2z^{(i)}\right)\right)\odot\left(1-y^2\right)\right)\sum_{m=1}^{D}
\left(\sum_{n=1}^{d}W_1^{(n,m)}W_2^{(i,n)}\left(1-(y^{(n)})^2\right)\right)^2\\
&~~~~~+ \sum_{i=1}^{D}\left(1-(z^{(i)})^2\right)^2\sum_{m=1}^{D}2
\left(\sum_{n=1}^{d}W_1^{(n,m)}W_2^{(i,n)}\left(1-(y^{(n)})^2\right)\right)\\
&~~~~~\times\left(\sum_{n=1}^{d}
\left(W_1^{(n,m)}W_2^{(i,n)}\left(-2y^{(n)}\right)\left(1-(y^{(n)})^2\right)\right)1_n^{D}\right)\\
&=-4\left(W_2'\odot \left(O^{d}\left(\left(1-z^2\right)^2\left(z\right)\right)'\right)\odot \left(\left(1-y^2\right)\left(O^{D}\right)'\right)\odot \left(O^{d}sum\left(\left(W_1'\left(W_2'\odot\left(1-y^2\right)(O^{D})'\right)\right)^2\right)\right)\right)O^{D}~~~\\
&~~~~~-4\left(sum\left(sum\left(\left(\left(O^{D}\left(\left(1-z^2\right)^2\right)'\right)\odot\left(W_1'\left(W_2'\odot\left(1-y^2\right)\left(O^{D}\right)'\right)\right)\right)\ominus \left(W_1\otimes W_2\right)\right)\right)\right)'\\
&~~~~~\odot\left(y-y^3\right)
\end{split}
\end{equation*}

\begin{equation*}
\scriptsize
\begin{split}
\frac{\partial ||J||_F^2}{\partial W_2}&=\sum_{i=1}^{D}\frac{\partial\left(1-(z^{(i)})^2\right)^2}{\partial W_2}\sum_{m=1}^{D}
\left(\sum_{n=1}^{d}W_1^{(n,m)}W_2^{(i,n)}\left(1-(y^{(n)})^2\right)\right)^2~~~~~~~~~~~~~~~~~~~~~~~~~~~~~~~~~~~~~~~~~~~~~~~~~~~~~~~~~~~~~~~~~~~~~~~~~\\
&~~~~~+\sum_{i=1}^{D}\left(1-(z^{(i)})^2\right)^2\sum_{m=1}^{D}\frac{\partial
\left(\sum_{n=1}^{d}W_1^{(n,m)}W_2^{(i,n)}\left(1-(y^{(n)})^2\right)\right)^2}{\partial W_2}\\
&=\sum_{i=1}^{D}1_i^{D}2(1-(z^{(i)})^2)^2(-2z^{(i)})y'\sum_{m=1}^{D}
\left(\sum_{n=1}^{d}W_1^{(n,m)}W_2^{(i,n)}\left(1-(y^{(n)})^2\right)\right)^2\\
&~~~~~+ \sum_{i=1}^{D}\left(1-(z^{(i)})^2\right)^2\sum_{m=1}^{D}2
\left(\sum_{n=1}^{d}W_1^{(n,m)}W_2^{(i,n)}\left(1-(y^{(n)})^2\right)\right)\left(1_i^{D}\left(W_1^{(mc)}\odot \left(1-y^2\right)\right)'\right)\\
&=-4\left(\left((1-z^2)^2z\right)\odot \left(sum\left(\left(W_1'\left(W_2'\odot\left(1-y^2\right)(O^{D})'\right)\right)^2\right)\right)'\right)y'\\
&~~~~~+2\left(\left(1-z^2\right)^2(O^{d})'\right)\odot \left(\left(W_1'\left(W_2'\odot \left(1-y^2\right)(O^{D})'\right)\right)'\left(W_1\odot \left(1-y^2\right)(O^{D})'\right)'\right)
\end{split}
\end{equation*}

\begin{equation*}
\scriptsize
\begin{split}
\frac{\partial ||J||_F^2}{\partial b_2}&=\sum_{i=1}^{D}\frac{\partial\left(1-(z^{(i)})^2\right)^2}{\partial b_2}\sum_{m=1}^{D}
\left(\sum_{n=1}^{d}W_1^{(n,m)}W_2^{(i,n)}\left(1-(y^{(n)})^2\right)\right)^2~~~~~~~~~~~~~~~~~~~~~~~~~~~~~~~~~~~~~~~~~~~~~~~~~~~~~~~~~~~~~~~~~~~~~~~~~\\
&=\sum_{i=1}^{D}2(1-(z^{(i)})^2)^2(-2z^{(i)})1_i^{D}\sum_{m=1}^{D}
\left(\sum_{n=1}^{d}W_1^{(n,m)}W_2^{(i,n)}\left(1-(y^{(n)})^2\right)\right)^2\\
&=-4\left((1-z^2)^2z\right)\odot \left(sum\left(\left(W_1'\left(W_2'\odot\left(1-y^2\right)(O^{D})'\right)\right)^2\right)\right)'
\end{split}
\end{equation*}

\subsubsection{Gradients for $Tr(JA')$}
\begin{equation*}
\scriptsize
\begin{split}
Tr(JA')&=Tr(\frac{\partial(z)}{\partial x}A')=\sum_{i=1}^{D}\left(1-(z^{(i)})^2\right)
\sum_{m=1}^{D}a_{mi}\left(\sum_{n=1}^{d}W_1^{(n,m)}W_2^{(i,n)}\left(1-(y^{(n)})^2\right)\right)~~~~~~~~~~~~~~~~~~~~~~~~~~~~~~~~~~~~~~~~~~~~~~~~~~
\end{split}
\end{equation*}

\begin{equation*}
\scriptsize
\begin{split}
\frac{\partial Tr(JA')}{\partial W_1}&=\sum_{i=1}^{D}\frac{\partial\left(1-(z^{(i)})^2\right)}{\partial W_1}
\sum_{m=1}^{D}a_{mi}\left(\sum_{n=1}^{d}W_1^{(n,m)}W_2^{(i,n)}\left(1-(y^{(n)})^2\right)\right)~~~~~~~~~~~~~~~~~~~~~~~~~~~~~~~~~~~~~~~~~~~~~~~~~~~~~~~~~~~~~~~~~~\\
&~~~~~+\sum_{i=1}^{D}\left(1-(z^{(i)})^2\right)\frac{\partial
\sum_{m=1}^{D}a_{mi}\left(\sum_{n=1}^{d}W_1^{(n,m)}W_2^{(i,n)}\left(1-(y^{(n)})^2\right)\right)}{\partial W_1}\\
&=\sum_{i=1}^{D}\left(\left(\left(W_2^{(i)}\right)'\left(1-(z^{(i)})^2\right)\left(-2z^{(i)}\right)\right)\odot\left(1-y^2\right)\right)x'
\sum_{m=1}^{D}a_{mi}\left(\sum_{n=1}^{d}W_1^{(n,m)}W_2^{(i,n)}\left(1-(y^{(n)})^2\right)\right)\\
&~~~~~+ \sum_{m=1}^{D}a_{mi}\left(\sum_{n=1}^{d}\delta_{n,m}^{W_1}W_2^{(i,n)}\left(1-(y^{(n)})^2\right)
+1_n^{d}\left(W_1^{(n,m)}W_2^{(i,n)}\left(-2y^{(n)}\right)\left(1-(y^{(n)})^2\right)\right)x'\right)\\
&~~~~~\sum_{i=1}^{D}\left(1-(z^{(i)})^2\right) \\
&=-2\left(W_2'\odot O^{d}\left(\left(z-z^3\right)\right)'\odot \left(1-y^2\right)O^{D}\odot O^{d}sum\left(\left(W_1'\left(W_2'\odot\left(1-y^2\right)(O^{D})'\right)\right)\odot A\right)\right)O^{D}x'\\
&~~~~~+\left(\left(O^{d}\left(1-z^2\right)'\right)\odot\left(W_2'\odot \left(\left(1-y^2\right)(O^{D})'\right)\right)\right)A'\\
&~~~~~-2\left(sum\left(sum\left(\left(\left(O^{D}\left(1-z^2\right)\right)\odot A\right)\ominus \left(W_1\otimes W_2\right)\right)\right)\right)'\odot\left(y-y^3\right)x'
\end{split}
\end{equation*}

\begin{equation*}
\scriptsize
\begin{split}
\frac{\partial Tr(JA')}{\partial b_1}&=\sum_{i=1}^{D}\frac{\partial\left(1-(z^{(i)})^2\right)}{\partial b_1}
\sum_{m=1}^{D}a_{mi}\left(\sum_{n=1}^{d}W_1^{(n,m)}W_2^{(i,n)}\left(1-(y^{(n)})^2\right)\right)~~~~~~~~~~~~~~~~~~~~~~~~~~~~~~~~~~~~~~~~~~~~~~~~~~~~~~~~~~~~~~~~~~\\
&~~~~~+\sum_{i=1}^{D}\left(1-(z^{(i)})^2\right)\frac{\partial
\sum_{m=1}^{D}a_{mi}\left(\sum_{n=1}^{d}W_1^{(n,m)}W_2^{(i,n)}\left(1-(y^{(n)})^2\right)\right)}{\partial b_1}\\
&=\sum_{i=1}^{D}\left(\left(\left(W_2^{(i)}\right)'\left(1-(z^{(i)})^2\right)\left(-2z^{(i)}\right)\right)\odot\left(1-y^2\right)\right)
\sum_{m=1}^{D}a_{mi}\left(\sum_{n=1}^{d}W_1^{(n,m)}W_2^{(i,n)}\left(1-(y^{(n)})^2\right)\right)\\
&~~~~~+ \sum_{i=1}^{D}\left(1-(z^{(i)})^2\right)\sum_{m=1}^{D}a_{mi}\left(1_n^{d}\left(W_1^{(n,m)}W_2^{(i,n)}\left(-2y^{(n)}\right)\left(1-(y^{(n)})^2\right)\right)\right)\\
&=-2\left(W_2'\odot O^{d}\left(\left(z-z^3\right)\right)'\odot \left(1-y^2\right)O^{D}\odot O^{d}sum\left(\left(W_1'\left(W_2'\odot\left(1-y^2\right)(O^{D})'\right)\right)\odot A\right)\right)O^{D}\\
&~~~~~-2\left(sum\left(sum\left(\left(\left(O^{D}\left(1-z^2\right)\right)\odot A\right)\ominus \left(W_1\otimes W_2\right)\right)\right)\right)'\odot\left(y-y^3\right)
\end{split}
\end{equation*}

\begin{equation*}
\scriptsize
\begin{split}
\frac{\partial Tr(JA')}{\partial W_2}&=\sum_{i=1}^{D}\frac{\partial\left(1-(z^{(i)})^2\right)}{\partial W_2}
\sum_{m=1}^{D}a_{mi}\left(\sum_{n=1}^{d}W_1^{(n,m)}W_2^{(i,n)}\left(1-(y^{(n)})^2\right)\right)~~~~~~~~~~~~~~~~~~~~~~~~~~~~~~~~~~~~~~~~~~~~~~~~~~~~~~~~~~~~~~~~~~~\\
&~~~~~+\sum_{i=1}^{D}\left(1-(z^{(i)})^2\right)\frac{\partial
\sum_{m=1}^{D}a_{mi}\left(\sum_{n=1}^{d}W_1^{(n,m)}W_2^{(i,n)}\left(1-(y^{(n)})^2\right)\right)}{\partial W_2}\\
&=\sum_{i=1}^{D}1_i^{D}(1-(z^{(i)})^2)(-2z^{(i)})y'
\sum_{m=1}^{D}a_{mi}\left(\sum_{n=1}^{d}W_1^{(n,m)}W_2^{(i,n)}\left(1-(y^{(n)})^2\right)\right)\\
&~~~~~+ \sum_{i=1}^{D}\left(1-(z^{(i)})^2\right)\sum_{m=1}^{D}a_{mi}\left(1_i^{D}\left(W_1^{(mc)}\odot \left(1-y^2\right)\right)'\right)
\\
&= -2\left(\left(z-z^3\right)\odot \left(sum\left(\left(W_1'\left(W_2'\odot\left(1-y^2\right)(O^{D})'\right)\right)\odot A\right)\right)'\right)y'\\
&~~~~~+\left(\left(1-z^2\right)(O^{d})'\right)\odot \left(\left(\left(W_1\odot \left(1-y^2\right)(O^{D})'\right)A\right)'\right)
\end{split}
\end{equation*}

\begin{equation*}
\scriptsize
\begin{split}
\frac{\partial Tr(JA')}{\partial b_2}&=\sum_{i=1}^{D}\frac{\partial\left(1-(z^{(i)})^2\right)}{\partial b_2}
\sum_{m=1}^{D}a_{mi}\left(\sum_{n=1}^{d}W_1^{(n,m)}W_2^{(i,n)}\left(1-(y^{(n)})^2\right)\right)~~~~~~~~~~~~~~~~~~~~~~~~~~~~~~~~~~~~~~~~~~~~~~~~~~~~~~~~~~~~~~~~~~~\\
&=\sum_{i=1}^{D}1_i^{D}(1-(z^{(i)})^2)(-2z^{(i)})
\sum_{m=1}^{D}a_{mi}\left(\sum_{n=1}^{d}W_1^{(n,m)}W_2^{(i,n)}\left(1-(y^{(n)})^2\right)\right)\\
&=-2\left(\left(z-z^3\right)\odot \left(sum\left(\left(W_1'\left(W_2'\odot\left(1-y^2\right)(O^{D})'\right)\right)\odot A\right)\right)\right)'
\end{split}
\end{equation*}

\end{document}